%% file: main.tex
\documentclass[11pt]{article}
\usepackage[utf8]{inputenc}
\usepackage[a4paper, total={6in, 8.2in}]{geometry}
\usepackage{amsmath, dsfont}
\usepackage{mathtools}
\usepackage{amsthm, amsmath}
\usepackage{amssymb}
\usepackage[english]{ babel}
\usepackage{dirtytalk}
\usepackage{graphicx}
\usepackage{authblk}
\usepackage[hyphens]{url}
\usepackage{hyperref}
\usepackage{caption}
\usepackage{subcaption}
\usepackage{cases}
\usepackage{xcolor} %%% temporary
\usepackage{bm}
\usepackage{tikz}
\usetikzlibrary{shapes,arrows.meta,positioning}

\usepackage{tabularx}
\usepackage{blkarray}
\usepackage{enumerate}
\usepackage{comment}
\usepackage{algorithm}
\usepackage{algorithmic}
\usepackage{float}
\usepackage{booktabs}
\usepackage{ragged2e}

\usepackage{appendix}
\usepackage{multirow}

\newcommand{\lr}[1]{\left(#1\right)}

\usepackage{amsmath}    % Advanced mathematical features and environments
\usepackage{amsthm}     % Theorem-like environments (theorems, lemmas, etc.)
\usepackage{amssymb}    % Additional mathematical symbols
\DeclareMathAlphabet{\mathcal}{OMS}{cmsy}{m}{n} % Custom math alphabet for calligraphic letters
\usepackage{mathtools} % Additional tools for mathematical typesetting
\usepackage{extarrows} % Extended arrows for math
\usepackage{stmaryrd} % Additional math symbols
\usepackage{arydshln} % Dashed lines in arrays/matrices
\usepackage{dsfont} % For indicator function

\usepackage[bottom]{footmisc}
\title{Analyzing Uncertainty Quantification in Statistical
and Deep Learning Models for Probabilistic
Electricity Price Forecasting}

\author[1]{Andreas Lebedev}
\author[2]{Abhinav Das}
\author[3]{Sven Pappert}
\author[4]{Stephan Schl\"uter}
\affil[1]{Institute of Energy Engineering and Energy Economics, Ulm University of Applied Sciences, 
Prittwitzstrasse 10, 89075 Ulm, Germany }
\affil[2]{Faculty of Mathematics and Economics, Ulm University, Helmholtzstrasse 20, 89081 Ulm, Germany }
\affil[3]{Faculty of Statistics, TU Dortmund University, Vogelpothsweg 78, 44227 Dortmund, Germany }
\affil[4]{Institute of Energy Engineering and Energy Economics, Ulm University of Applied Sciences, 
Prittwitzstrasse 10, 89075 Ulm, Germany }

\date{}

\begin{document}
\maketitle
\begin{abstract}
Precise probabilistic forecasts are fundamental for energy risk management, and there is a wide range of both statistical and machine learning models for this purpose. 
Inherent to these probabilistic models is some form of uncertainty quantification.
However, most models do not capture the full extent of uncertainty, which arises not only from the data itself but also from model and distributional choices.
In this study, we examine uncertainty quantification in state-of-the-art statistical and deep learning probabilistic forecasting models for electricity price forecasting in the German market.
In particular, we consider deep distributional neural networks (DDNNs) and augment them with an ensemble approach, Monte Carlo (MC) dropout, and conformal prediction to account for model uncertainty.
Additionally, we consider the LASSO-estimated autoregressive (LEAR) approach combined with quantile regression averaging (QRA), generalized autoregressive conditional heteroskedasticity (GARCH), and conformal prediction.
Across a range of performance metrics, we find that the LEAR-based models perform well in terms of probabilistic forecasting, irrespective of the uncertainty quantification method.
Furthermore, we find that DDNNs benefit from incorporating both data and model uncertainty, improving both point and probabilistic forecasting.
Uncertainty itself appears to be best captured by the models using conformal prediction.
Overall, our extensive study shows that all models under consideration perform competitively. 
However, their relative performance depends on the choice of metrics for point and probabilistic forecasting.
\hfill\newline\newline
Keywords: Bayesian methods, deep learning, electricity price forecasting, ensemble learning, interval forecasting, machine learning, Monte Carlo methods, power markets, probabilistic forecasting, uncertainty quantification
\vfill
\hfill\newline
Preprint Submitted to IEEE Access
\end{abstract}
\newpage
\section{Introduction}
\label{sec:Intro}

The energy industry has been in turmoil lately. 
On the positive side, we see a considerable increase in renewable energy, leading to a reduced carbon footprint of power production. 
However, this comes with a drawback, as renewable power can be highly fluctuating, which leads to increased production risk and larger price volatility in turn. 
This volatility is further amplified by various geopolitical developments in 2025, such as the ongoing war in Ukraine. 
Risk management for power utilities, energy-related companies, and governments is becoming not only important but essential. 
This involves reliable forecasts. In risk management, probabilistic/interval forecasts are of particular interest, as point forecasts ignore information about uncertainty arising from model choice, data, or parameter calibration.

In the electricity price forecasting literature, several authors have focused on developing (probabilistic) forecasting models for power prices. 
See, e.g., \cite{Weron2014ElectricityFuture}, \cite{Lago2021ForecastingBenchmark}, and \cite{Lago2018ForecastingAlgorithms} for overviews. 
For general overviews of probabilistic/interval forecasting, see \cite{tay2000density} and \cite{wang2024interval}.

%Lately, statistical approaches are challenged by machine learning-based approaches that involve distributional deep neural networks (DDNNs), see \cite{marcjasz2023distributional}. 

Despite various refined models, existing research often addresses data-related uncertainty, whereas uncertainty arising from model choice and calibration is often overlooked (see Section \ref{sec:uncertainty} for more details). 
Including this uncertainty, e.g., via ensemble or Bayesian approaches, considerably increases complexity and computational challenges, and it remains to be verified whether the additional effort pays off. 
So far, to the authors' best knowledge, there has been no deeper analysis in electricity price forecasting with regard to this question. 
Hence, we propose to compare probabilistic forecasts and prediction intervals (PIs) generated by distributional deep learning and statistical models in combination with various uncertainty quantification methods.
In addition, we discuss which evaluation measure is adequate here.
For illustration and testing purposes, we use historical energy price data from the German market. 
These data not only show high volatility but also multiple structural breaks in the past and are therefore adequate for thoroughly testing methods for uncertainty quantification and comparison with forecasts from other models. 
%In our forecasting study, we find that models from the related literature do not take model uncertainty into consideration.
In this work, we demonstrate that accounting for both data and model uncertainty, particularly for deep learning models, yields significant benefits. As shown in Section \ref{sec:CaseStudy}, this approach leads to more reliable PIs as well as improved point forecasts. We also find that statistical models are highly competitive, even when compared to complex machine learning models.

The article is structured as follows: In Section \ref{sec:theory}, we discuss state-of-the-art approaches for (probabilistic/interval) electricity price forecasting from both machine learning and statistics, commenting on their approaches to quantifying uncertainty and motivating additional approaches to uncertainty quantification.  
%We also introduce our model and comment on uncertainty in general, as this motivates our approach.  
In Section \ref{sec:Prices}, the dataset for the case study is introduced. The forecasting study is presented in Section \ref{sec:CaseStudy}, where we discuss how models are calibrated and trained (including hyperparameter choices), and compare results using chosen metrics such as the prediction interval coverage probability (PICP) and the continuous ranked probability score (CRPS) (see \cite{gneiting2007strictly}). Before we conclude the article in Section \ref{sec:conclusion}, we discuss our results in Section \ref{sec:discussion}.

\section{A Revision of Concepts for Forecasting and Uncertainty Quantification}
\label{sec:theory}

There are various approaches for generating probabilistic and interval forecasts. 
First, there are parametric distribution-based approaches that rely on certain assumptions about the underlying probability distribution of the data. 
The PIs are constructed by estimating the parameters of the (conditional) distribution and using the corresponding quantiles, as in distributional deep neural networks (DDNNs) \cite{Marcjasz2023DistributionalForecasting}. 
Then, some methods directly learn specific aspects of a forecast, e.g., direct quantile learning methods such as quantile regression \cite{Nowotarski2018RecentForecasting}, which do not require distributional assumptions. 
Finally, there are model-agnostic approaches such as bootstrapping \cite{Chen2012ElectricityBootstrapping}, historical simulation \cite{Lipiecki2024PostprocessingRegression}, or conformal prediction \cite{Lipiecki2024PostprocessingRegression}, which do not rely on any specific underlying distribution or regression model. 
These approaches are applied to already trained models.

In Section \ref{sec:uncertainty}, we first review concepts of uncertainty quantification. 
Then we proceed to introduce state-of-the-art statistical and machine learning models for probabilistic electricity price forecasting and discuss them in the context of uncertainty.

\subsection{A Few Notes on Uncertainty}
\label{sec:uncertainty}

Forecasting intervals are an attempt to quantify and estimate the uncertainty connected to the behavior of a stochastic process. However, uncertainty itself needs to be taken into account when designing or choosing a model for the generation of PIs.

Generally speaking, uncertainty can be categorized into two types: aleatoric uncertainty, which describes data-related uncertainty \cite{Gawlikowski2023ANetworks}, and epistemic uncertainty, which relates to model-specific uncertainties \cite{He2025ALearning, Abdar2021AChallenges}. 
Aleatoric uncertainty, also known as data uncertainty, arises from the variability inherent in the input data and cannot be reduced, not even by collecting more data. 
Sources of aleatoric uncertainty include measurement errors or the use of inherently uncertain input features, such as energy demand estimates or weather forecasts. 
Epistemic uncertainty, on the other hand, arises from uncertainties related to the choice of model features, such as the general architecture (e.g., using a linear model to capture nonlinear relationships), the parameters, and the distributional properties. 
Especially when dealing with nonstationary processes and potential structural breaks, we must address domain-shift uncertainty, which describes potential differences in the distributional properties of calibration and test data. 
Unlike aleatoric uncertainty, epistemic uncertainty can be reduced. 
Key strategies for minimizing it include selecting a suitable model architecture, employing an effective optimization and training process, and using a well-distributed and representative training dataset. 
Ultimately, the resulting overall uncertainty related to the prediction for a specific sample is called predictive uncertainty.

\subsection{Statistical Approaches}
\label{sec:theoryStatistics}
The standard approach to statistical interval forecasting is to assume an underlying distribution and obtain the interval by extracting the corresponding quantiles. 
The simplest way to model a distribution is the naive model. The model naively uses a previous days price as mean forecast.
%It uses the previous day's price to forecast prices for Tuesday through Friday, and it uses prices from the same day of the previous week to predict prices for Monday, Saturday, and Sunday. 
To estimate price distributions, historical simulation may be applied, relying on the widely used Gaussian distribution. 
This involves fitting distribution parameters to the forecast errors from a selected dataset. 
We refer to these models as Naive-HS$_{train}$ and Naive-HS$_{val}$, depending on whether the training or validation set was used.

With regard to the distribution's volatility parameters, there are more complex approaches, such as the generalized autoregressive conditional heteroskedasticity (GARCH) model (see \cite{bollerslev1986generalized}), which models the volatility parametrically, using an autoregressive model for it. 
Other parametric approaches such as copulas, see e.g. \cite{pappert2023forecasting}, might be used as well but are beyond the scope of this paper. The uncertainty may also be captured by ensemble methods or conformal prediction.
%Authors such as Berrisch and Ziel \cite{berrisch2022distributional} or Pappert and Arsova \cite{pappert2023forecasting}, for example, suggest using Copulas, which are a generalized distribution approach.

We consider three statistical models for probabilistic forecasting, namely the LEAR-QRA model by \cite{Lago2021ForecastingBenchmark}, Sect.~4.2 (see also \cite{Marcjasz2023DistributionalForecasting}, Sect.~4.3.2), the LEAR model with GARCH residuals, LEAR-GARCH, and the LEAR model with conformal prediction, LEAR-CP.
LEAR stands for LASSO-estimated autoregressive, and QRA for quantile regression averaging. 
In the following, both benchmark models are briefly described. 
We start by introducing the LEAR model for mean forecasting, which is the aspect both models share. 
Afterwards, the distributional modeling of the two approaches is described.

The LEAR model, originally introduced by \cite{uniejewski2016automated}, assumes an autoregressive model with external regressors for each product individually and then estimates the coefficients using a LASSO penalty (see \cite{tibshirani1996regression} for the LASSO penalty). For each product, the model may be written as
\begin{align*}
    y_t = \beta_0 + Y_t\beta + X_t\gamma + \varepsilon_t
\end{align*}
where $\{y_t\}_{t\in\mathbb{Z}}$ is the time series of interest, $Y = (y_{t-1},\hdots,y_{t-p})$ contains past values of $\{y_t\}$, and $X_t = (x_{1,t-1},\hdots,x_{1,t-p_1},\hdots,x_{r,t-1},\hdots,x_{r,t-p_r})$ contains the past values of the external regressors. 
The parameters are estimated by minimizing the LASSO-augmented sum of squares. 
If the penalty parameter is chosen correctly, the estimator will select only the relevant past lags of $y_t$ and the external regressors. This enhances precision and decreases the variability of the mean predictions.
Since LEAR produces only mean forecasts, there is a need to extend the forecasting procedure to yield probabilistic forecasts. 
The variance is of particular interest in this regard. 
We consider two options: 1) applying quantile regression averaging to different mean predictions, i.e., LEAR-QRA, as proposed by \cite{Lago2021ForecastingBenchmark}; and 2) fitting a GARCH model on the residuals of the LEAR model to obtain predictions for the conditional variance.

For the LEAR-QRA model, different mean predictions are necessary.
We follow the procedure of \cite{Lago2021ForecastingBenchmark} and train the LEAR model for different horizons to obtain different mean forecasts.
In total, $m$ mean forecasts are obtained (in the application, we set $m=4$).
Then, a quantile regression with these mean forecasts as regressors is fitted for each percentile.
To obtain probabilistic forecasts for an observation from the test data, first, the LEAR models are re-fitted for the $m$ horizons, and the respective mean predictions are obtained.
Second, the mean predictions are used in conjunction with the fitted quantile regression coefficients to obtain predictions for each percentile.
The collection of the percentiles constitutes the probabilistic forecast.

The LEAR-GARCH model follows a different approach.
Instead of relying on ensemble-based estimators for the probabilistic forecast, the conditional variance is predicted by a GARCH model.
Assuming an underlying conditional distribution, the conditional mean (given by the LEAR prediction) and the conditional variance (given by the GARCH prediction) constitute a probabilistic forecast (in the application, we assume that the conditional distribution is the normal distribution). Last, in the LEAR-CP model, conformal prediction is applied to the mean-forecasts of the LEAR model.

While sharing a basic building block (the LEAR mean modeling), these approaches are fundamentally different. The GARCH approach quantifies uncertainty explicitly by modeling the conditional variance, whereas the LEAR-QRA approach captures uncertainty implicitly using an ensemble of predictors trained on different horizons. In particular, short-term training windows can account for recent increases in variance, as discussed in \cite{Lago2021ForecastingBenchmark}. Then the conformal prediction approach offers a third, and different approach to uncertainty quantification, as is explained in the next section.

\subsection{Deep Learning-Based Concepts}
\label{sec:theoryML}

As machine learning models are widely used today, we refrain from providing a detailed introduction to deep learning and machine learning models in general.
If the reader is not familiar with such concepts, we refer to \cite{aggarwal2018}.

Lago et al. \cite{Lago2021ForecastingBenchmark} show that simple two-layer feed-forward neural networks are among the most effective deep learning models for point forecasting.
Building on this, Marcjasz et al. \cite{Marcjasz2023DistributionalForecasting} proposed the distributional deep neural network (DDNN), a probabilistic extension that estimates the parameters of the underlying price distribution rather than predicting electricity prices directly.
The model is trained by minimizing the negative log-likelihood loss: 
\begin{equation*}\label{eq:loss_func}
L_{NLL} = - \sum_{i=1}^{N}\log p(y_{i}\mid x_{i};\theta),
\end{equation*}
where $x_{i}$ and $y_{i}$ are the inputs and observed targets, and $p(y_{i}\mid x_{i};\theta)$ the probability assigned to $y_{i}$ by the distribution predicted by the network with parameters $\theta$.
The output distribution captures data uncertainty in the predictions \cite{Gawlikowski2023ANetworks}.
To additionally capture model uncertainty, various approaches can be used, including ensemble methods, Bayesian approaches such as Monte Carlo Dropout, and single deterministic methods such as Deep Evidential Regression \cite{Gawlikowski2023ANetworks}.

\subsubsection{Ensembles}
% Ensembles
The idea of ensemble methods is to combine different models, i.e., their forecasts, to generate improved predictions, which come at a cost, as multiple models must be calibrated and maintained.
% With the increasing application of machine learning tools in time series analysis, ensemble forecasts have also been introduced to this field \cite{Dietterich2000EnsembleLearning}.
% Here, ensembles are often not combinations of different methods but rather several neural networks \cite{Lakshminarayanan2017SimpleEnsembles}.
% With regard to neural networks, ensemble models offer a way to evaluate uncertainty, as explained in \cite{Abdar2021AChallenges} and \cite{Mendes-Moreira2012EnsembleSurvey}. 

However, ensembles also provide an intuitive way to measure model uncertainty by capturing the variation in predictions across the models.
For NNs, it is essential to maximize the diversity of the network's behavior to obtain reliable uncertainty estimates (while maintaining good predictive performance).
This can be done by training different architectures, training on different subsets of the original dataset, or varying the initial parameter set \cite{Mendes-Moreira2012EnsembleSurvey}.
% Here, we construct an ensemble model that consists of $N > 0$ DDNNs and their outputs.
Lakshminarayanan et al. \cite{Lakshminarayanan2017SimpleEnsembles} found that using different random initializations of the NN weights and random shuffling suffices to create network diversity.
To get the predictive distribution, one has to aggregate the ensemble models into a weighted mixture model:
% The respective PDF is constructed as a linear combination
\begin{eqnarray*}
p(y\mid x;\theta)& = &\sum_{j=1}^{M}w_{j}\, p(y\mid x;\theta_j)  \\
\end{eqnarray*}
using nonnegative weights $w_1,\ldots,w_M$ that add up to one.
% The cumulative distribution function is defined accordingly, with $\theta=(\mu_{1},...,\mu_{N},\sigma_{1},...,\sigma_{N})$ and weights $w_{i},i=1,...,N$ \cite{Lago2018ForecastingAlgorithms}.
To calculate the quantile $Q(p;\theta)=F^{-1}(p;\theta)$ for a given probability $p\in[0,1],$ we solve the following equation: % \cite{Marcjasz2018SelectionForecasting}:
\begin{equation}
F(x;\theta)-p=0 \label{eq:4.18}
\end{equation}
numerically. 
The initial guess for the solution we set to $x_{0}= \sum_{i=1}^{M}w_{j}F_j^{-1}(p).$

% \subsection{Bayes Logic for Interval Forecasts} 
% \label{sec:bayesIntervalForecasts}
% MCD
\subsubsection{Monte Carlo Dropout}
While ensemble models train several NNs to capture model uncertainty, Bayesian logic suggests an alternative approach by interpreting individual weights as probabilistic variables.
The Monte Carlo (MC) dropout model is a widely used example \cite{Gal2016DropoutLearning}.
It differs from the DDNN by including dropout layers after each hidden layer.
These dropout layers remain active during both training and inference.
As a result, the same input can produce different outputs in multiple forward passes through the network.
During inference, we perform $N$ forward passes and aggregate the results using a uniformly weighted mixture model.
Quantiles are then obtained by solving Equation \ref{eq:4.18}, equivalent to the ensemble approach.
% The resulting NLL loss function using a Gaussian mixture distribution reads as follows:
% \begin{equation*}
% L_{NLL,GM}=-\sum_{i=1}^{N_{d,i}}\log\left(\sum_{j=1}^{N}\frac{1}{\sqrt{2\pi\hat{\sigma}_{i,j}^{2}}}\exp\left(-\frac{(y_{i}-\hat{\mu}_{i,j})^{2}}{2\hat{\sigma}_{i,j}^{2}}\right)\right) \label{eq:4.19}
% \end{equation*}
% with predicted outputs $\hat{\mu}_{i,j}$ and $\hat{\sigma}_{i,j}$ of the NN for each forward pass. 
% As with the DDNN model, we predict $\log \sigma^{2}$ instead of $\sigma^{2}$ directly.

% EvDNN
\subsubsection{Deep Evidential Regression}
Deep evidential regression (DER) is a single deterministic method, introduced by Amini et al. \cite{Amini2020DeepRegression}, where one evidential deep neural network (EvDNN) predicts both the forecast and its uncertainty. In contrast to a Bayesian method, priors are placed directly over the parameters of the distribution instead of the network weights. If a Gaussian distribution is assumed, a Gaussian prior is placed on the mean and an inverse gamma prior on the variance:
\begin{align*}
\begin{split}
    y &\sim \mathcal{N}(\mu,\sigma^2), \\ 
    \text{with } \mu &\sim \mathcal{N}(\gamma,\sigma^2\nu^{-1}), \quad
    \sigma^2 \sim \Gamma^{-1}(\alpha,\beta),
\end{split}
\end{align*}
where $\gamma \in \mathbb{R}$, $\nu > 0$, $\alpha > 1$, and $\beta > 0$.
This leads to a normal-inverse gamma distribution for the joint prior distribution, assuming that the parameters factorize as $p(\mu,\sigma^2)=p(\mu)(\sigma^2)$:
\begin{equation*} \label{eq:EDL1}
    \begin{split}
    &p(\theta|m)\\
    &=\dfrac{\beta^\alpha \sqrt{\nu}}{\Gamma(\alpha)\sqrt{2\pi\sigma^2}} \lr{\dfrac{1}{\sigma^2}}^{\alpha+1} \exp\lr{-\dfrac{2\beta+\nu(\gamma-\mu)^2}{2\sigma^2}} \\&=NIG(\mu,\sigma^2;\gamma,\nu,\alpha,\beta)
    \end{split},
\end{equation*}
where we write $\theta=(\mu,\sigma^2)$ and $m=\gamma,\nu,\alpha,\beta$.
Integrating over $\theta$ yields the predictive distribution for $y$:
\begin{equation*} \label{eq:EDL2}
\begin{split}
 p(y|m)&=\int p(y|\theta)p(\theta|m) \, d\theta\\
&= St\lr{y;\gamma,\dfrac{\beta(1+\nu)}{\nu\alpha};2\alpha}.
\end{split}
\end{equation*}
This predictive distribution follows a Student’s t-distribution with location $\gamma$, scale $\tfrac{\beta(1+\nu)}{\nu\alpha}$, and $2\alpha$ degrees of freedom. The network is then trained by minimizing the NLL loss
\begin{equation*}
\begin{split}
    L_{DER,NLL} &= \sum_{i=1}^N St\lr{y_i;\hat{\gamma}_i,\dfrac{\hat{\beta}_i(1+\hat{\nu}_i)}{\hat{\nu}_i\hat{\alpha}_i};2\hat{\alpha}_i}\\
    &= \sum_{i=1}^N \dfrac{1}{2}\log \lr{\dfrac{\pi}{\hat{\nu}_i}} - \hat{\alpha}_i\log (\hat{\Omega}_i) \\
    &+ \lr{\hat{\alpha}_i+\dfrac{1}{2}}\log \lr{(y-\hat{\gamma}_i)^2\hat{\nu}_i + \hat{\Omega}_i} \\
    &+\log \lr{\dfrac{\Gamma(\hat{\alpha}_i)}{\Gamma \lr{\hat{\alpha}_i+\frac{1}{2}}}},
\end{split}
\end{equation*}
where $\hat{\Omega}_i = 2\hat{\beta}_i(1+\hat{\nu}_i)$, $\Gamma(\cdot)$ denotes the Gamma function, and $\hat{\gamma}_i$, $\hat{\nu}_i$, $\hat{\alpha}_i$, and $\hat{\beta}_i$ are the predicted parameters for each training sample $(x_i,y_i)$.
In addition to the NLL loss, an evidential regularizer is introduced to penalize predictions that are inaccurate while exhibiting low uncertainty. This term is defined as
\begin{equation*}
    L_{DER,R} = \sum_{i=1}^N |y_i - \hat{\gamma}_i|\cdot(2\hat{\nu}_i+\hat{\alpha}_i).
\end{equation*}
The overall loss function for the DER method is the sum of the NLL loss and the regularizer, weighted by a hyperparameter $\lambda_{DER} > 0$:
\begin{equation*}\label{eq:DER}
    L_{DER} = L_{DER,NLL} + \lambda_{DER}\, L_{DER,R}.
\end{equation*}

\subsection{Model-Agnostic Approaches}

Model-agnostic approaches do not require assumptions about a specific probability distribution or regression model. 
Instead, they can be applied on top of any forecasting model. Historical simulations are examples, as described in Section \ref{sec:theoryStatistics}.

Another example is Conformal Prediction \cite{Lipiecki2024PostprocessingRegression}. 
Given a trained model on historical data, PIs are constructed using nonconformity scores, computed from the residuals of a distinct calibration set for each hour of the day.
For each day $d$ and hour $h$, the absolute residual as the nonconformity score is computed as,
\begin{equation*}
s_{d,h} = |\hat{y}_{d,h} - y_{d,h}|
\end{equation*}
where $\hat{y}_{d,h}$ is the point prediction and $y_{d,h}$ is the observed electricity price at hour $h$ of day $d$.

For a desired quantile $q$ and each hour $h$, we construct symmetric PIs as follows:
For the lower quantile ($q < 0.5$):
\begin{equation*}
\hat{q}_{d,h}^{(q)} = \hat{y}_{d,h} - Q_{1-2q}(S_{d,h})
\end{equation*}
For the upper quantile ($q \geq 0.5$):
\begin{equation*}
\hat{q}_{d,h}^{(q)} = \hat{y}_{d,h} + Q_{2q-1}(S_{d,h})
\end{equation*}

where $Q_{\alpha}(S_{d,h})$ denotes the $\alpha$-quantile of the hour-specific nonconformity scores
$S_{d,h} = \{s_{d-n_{cal},h}, s_{d-n_{cal}+1,h}, \ldots, s_{d-1,h}\}$.

\section{Electricity Data}
\label{sec:Prices}

All data were retrieved from the SMARD platform (smard.de), operated by the German Federal Network Agency. 
For forecasting power prices, we consider not only the price itself but also load forecasts and renewable energy generation forecasts. 
Daily seasonality is captured using weekday dummies. 
The dataset covers the period from 1 October 2018 to 30 November 2024. 
In general, the data quality is very high. 
However, entire days are regularly missing in the load forecast series.
% For all but load forecast data, we have observations that range from 1st October 2018 to 30 November 2024. In general, the data quality of all safe for load forecasts is very good. For the latter one, regularly full days are missing. 
The power prices we consider are hourly day-ahead prices with physical delivery, traded on the European Power Exchange (EPEX SPOT). 
The day-ahead auction is held daily, with price settlement taking place at noon for delivery on the following day \cite{Marcjasz2023DistributionalForecasting}.
% Commet: Consitent with \cite{Marcjasz2023DistributionalForecasting}.

% The power prices we consider are hourly (with physical delivery) that are traded day-ahead on the European Power Exchange EPEX SPOT. Price settlement is at noon the day before. Prices for Monday are jointly settled with Saturday and Sunday on Friday. The same holds for public holidays. 

\begin{figure}[ht!]
    \centering
    \includegraphics{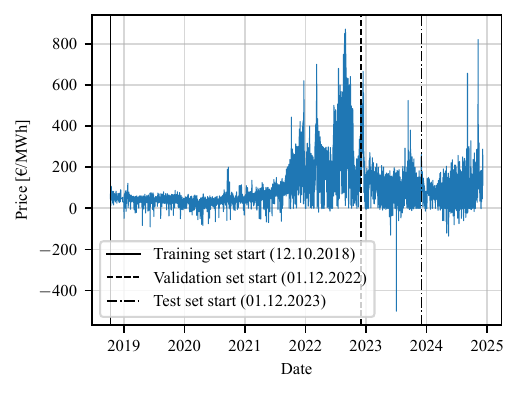}
    \caption{Day-ahead electricity prices from 12.10.2018 to 30.11.2024.}
    \label{fig:prices_dataset}
\end{figure}
An overview of the dataset is provided in Fig.~\ref{fig:prices_dataset}, which shows several noteworthy characteristics. 
Negative prices appear regularly. 
However, the period between mid-2021 and December 2022 stands out as an exception, with almost no negative prices, comparatively high volatility, and considerably large price spikes. 
Price spikes can also be observed in other years and are due to the physical delivery of electricity, combined with the fact that, up to now, it cannot be stored in large quantities in Germany.
As a consequence, in November 2024, when Germany experienced a period of low wind and limited solar generation, together with insufficient fossil power, this led to a price surge of more than 800 EUR per MWh. 
In general, price properties vary considerably over time, which is why we refrain from computing fundamental statistical measures such as skewness, variance, or kurtosis. 
Fig.~\ref{fig:prices_dataset} further supports the assumption of time-varying volatility.
\begin{figure}[ht!]
    \centering
    \includegraphics{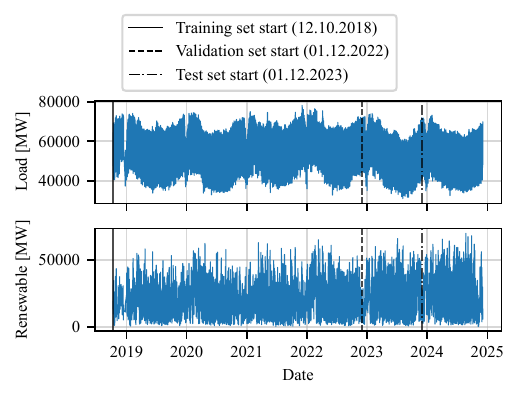}
    \caption{Load and renewable energy (wind + PV) forecasts from 12.10.2018 to 30.11.2024.}
    \label{fig:loadAndRenewables}
\end{figure}

\begin{figure}[ht!]
    \centering
    \includegraphics[width = 7cm, height = 5cm]{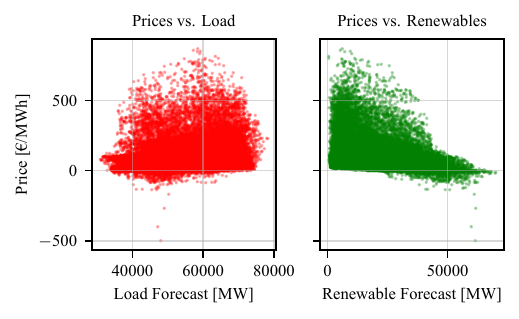}
    \caption{Load and renewable forecasts against day-ahead electricity prices from 12.10.2018 to 30.11.2024.}
    \label{fig:scatterPlots}
\end{figure}

\section{Case Study}
\label{sec:CaseStudy}
% Besides, the augmented Dickey-Fuller test rejects the stationarity assumption.

Both renewable energy generation and load forecasts are displayed in Fig.~\ref{fig:loadAndRenewables}. 
Load exhibits clear annual seasonality, with lower demand in summer than in winter. 
While overall load volumes remain relatively stable across years, the structure within individual years is changing, though not to the extent of power prices. 
Forecasted renewable generation is increasing over time, consistent with infrastructure efforts in Germany. 
At this level, no distinct seasonality is visible, which may be explained by the negative correlation between forecasted solar and wind generation.

% TODO (all): Isn't it the German date style?

Eventually, Fig.~\ref{fig:scatterPlots} presents scatter plots of load forecasts vs.\ power prices and renewable forecasts vs.\ power prices. 
Across the dataset, there is no clear dependence structure visible. 
However, this does not necessarily indicate the absence of interdependence but rather suggests a time-varying dependence structure.

To evaluate whether incorporating model uncertainty into deep learning models improves interval forecasts, we assess eight augmentations of the distributional deep neural network (DDNN) on the dataset described in Section \ref{sec:Prices}.
These include ensemble models of 5 and 10 DDNNs (Ens5 and Ens10), MC Dropout models with 10 or 30 forward passes (MCD10 and MCD30), and an evidential deep neural network (EvDNN).
Both ensembles and dropout models employ a Gaussian mixture aggregation scheme.
In addition, all point forecasts (excluding Ens5 and MCD10) are combined with conformal prediction (suffix "–CP").
The DDNN, which accounts only for data uncertainty, serves as the baseline.
For comparison, we also consider statistical benchmarks: LEAR-GARCH, LEAR-QRA, LEAR point forecasts with conformal prediction (LEAR-CP), and a naive approach combined with historical simulation based on the training and validation datasets (Naive-HS$_{train}$ and Naive-HS$_{val}$). 
In total, we evaluate 15 models.

% To test whether the deep learning approaches of uncertainty quantification improve interval forecasts we test ten different forecasting models on the data described in Section \ref{sec:Prices}, namely the distributional deep neural network (DDNN), an ensemble model consisting of 5 and one consisting of 10 networks (in short Ens5, Ens10), an evidential deep neural network (EvDNN) a LEAR-GARCH model (LEAR-GARCH), the LEAR quantile regression averaging model (LEAR-QRA), two MC dropout models with 10 or 30 forward passes (short MCD10 or MCD30), and eventually the naive approaches as described above (Naive-HS$_{train}$ and Naive-HS$_{val}$). We use the DDNN model as our baseline. For uncertainty quantification,  the ensemble method with a Gaussian mixture aggregation scheme is used. All deep learning models share the same architecture, consisting of two hidden layers with $1,024$ neurons each. The hidden layers use the ReLU activation function, whereas the output layer uses a linear activation function. Note that the suffix CP is added to models that are combined with conformal prediction. 
% \subsection{Preprocessing and Model Calibration}

\subsection{Input features and Preprocessing}
\label{subsec:ModelCalibrate}

The following $151$ features are used as inputs to our models:
\begin{itemize}
    \item Day-ahead prices from the past three days and from the same weekday of the previous week,
    \item Day-ahead forecasts of total load,
    \item Day-ahead forecasts of renewable energy generation,
    \item Weekday dummies, represented as binary vectors (e.g., [1, 0, 0, 0, 0, 0, 0] indicates Monday).
\end{itemize}

First, daylight saving time adjustments are handled by interpolating the missing hour during the spring shift with the average of the neighboring hours and resolving the duplicated hour during the autumn shift by averaging the two values. 
Next, we check for missing values. 
Out of a total of $2,253$ daily entries, $65$ contain missing data. 
These rows are removed, resulting in a final dataset that begins on 12 October 2018. 
Since the features have different scales, all $151$ input features and $24$ output values are standardized.
Finally, the dataset is split into training, validation, and test sets. 
The training set covers 12 October 2018 to 30 November 2022, the validation set spans 1 December 2022 to 30 November 2023, and the test set includes all days between 1 December 2023 and 30 November 2024.

\subsection{Model Calibration and Training}

%\subsubsection{Statistical Models}

%\subsubsection{Deep Learning Models}

All deep learning models share the same architecture, consisting of two hidden layers with 1,024 neurons each. 
The hidden layers use ReLU activation functions, while the output layer employs a linear activation.
All deep learning models use the "He" initialization for the hidden layers and the "Glorot" initialization for the output layer. 
During training, early stopping is applied with a patience of 100 epochs, and the maximum number of training epochs is set to 2000. 
The models are optimized using the ADAM optimizer with a batch size of 32. To reduce overfitting, all models employ L$_2$ regularization. 
Additionally, 10 independent runs are performed for each model, and the results are averaged. 
To ensure reproducibility, a fixed random seed is used for each training run.
For the MCD10 and MCD30 models, only a single forward pass is performed during training, while validation involves 10 and 30 forward passes, respectively.
To improve numerical stability during training, we follow the approach of \cite{Kendall2017WhatVision} and predict the logarithm of the variance, $\log \sigma^{2}$, instead of predicting the variance $\sigma^{2}$ directly.
For the EvDNN, we similarly predict $\log \nu$, $\log (\alpha - 1)$, and $\log \beta$ instead of $\nu$, $\alpha$, and $\beta$, and for the DDNN we predict $\log \sigma^{2}$ in place of $\sigma^{2}$.

For the LEAR model, we follow \cite{Lago2018ForecastingAlgorithms} and train for different forecasting horizons, specifically 56, 84, 1092, and 1461 days. 
We average the results to produce the final point forecast. 
Additionally, we employ a rolling window scheme. 
The QRA and GARCH approaches use the mean forecasts obtained from the LEAR model based on the validation set.
We fit 24 hourly GARCH models, each trained only on data from its corresponding hour.
For the naive model, we use the previous day's price as the mean forecast for days during Tuesday and Friday, and prices from the same day of the previous week to predict prices for Monday, Saturday, and Sunday. 

% \subsubsection{Conformal Prediction}
For conformal prediction, we use the validation samples as a calibration set, following \cite{Lipiecki2024PostprocessingRegression}. 
For each test day $d$ and hour $h$, we apply a rolling window of the most recent nonconformity scores specific to that hour, ensuring that only scores from the same hour across different days are used.

\subsection{Hyperparameter Optimization}

Before training, we performed hyperparameter optimization using the tree-structured Parzen estimator, a Bayesian optimization method implemented in the Optuna framework \cite{ns_optuna_2019}. 
For the deep learning models, we optimized the learning rate, regularization parameters, dropout rate, and the evidence regularization parameter, with their corresponding search ranges listed in Table \ref{tab:hp_opt}.
All hyperparameters, except the dropout rate, were sampled on a logarithmic scale. 
The optimization process consisted of 200 trials. 
The validation loss was used as the objective value for the tree-structured Parzen estimator algorithm. 
In each trial for the NN models, three runs were trained using the same hyperparameter values to reduce the effect of randomness, and the average validation loss across these three runs was then used as the objective value.

For the LEAR models, we optimized the LASSO penalty parameter using only the specification with the largest calibration window (1,461 days), and the resulting parameter was applied across all forecasting horizons.

\subsection{Error Metrics}
\label{sec:errorMetric}

To measure point prediction accuracy, both the mean absolute error (MAE) and the root mean square error (RMSE) are used, with
\begin{eqnarray*} %\label{em:mae}
    {MAE}  &=& \frac{1}{24 N_d} \sum_{d=1}^{N_d} \sum_{h=1}^{24} \left| p_{d,h} - \hat{p}_{d,h} \right|,\\
    {RMSE} &=& \sqrt{ \frac{1}{24 N_d} \sum_{d=1}^{N_d} \sum_{h=1}^{24} \left( p_{d,h} - \hat{p}_{d,h} \right)^2 },
\end{eqnarray*}
where $p_{d,h}$ and $\hat{p}_{d,h}$ denote the observed and predicted prices at day $d$ and hour $h$, respectively, and $N_d$ is the number of prediction days.

For interval forecasts, we use the prediction interval coverage probability (PICP), the mean absolute average coverage error (MAACE), the continuous ranked probability score (CRPS), and the mean prediction interval width (MPIW). 
The PICP is defined as
\begin{align*}
    PICP_{1-\alpha} &= \dfrac{1}{N_{d,h}}\sum_{i=1}^{N_{d,h}} I_{d,h}^{(1-\alpha)}, \\
    I_{d,h}^{(1-\alpha)} &= 
    \begin{cases}
        1 & \text{if } p_{d,h} \in [ L_{d,h}^{(1-\alpha)},\ U_{d,h}^{(1-\alpha)} ], \\
        0 & \text{otherwise},
    \end{cases}
\end{align*}
with $L_{d,h}^{(1-\alpha)}$ and $U_{d,h}^{(1-\alpha)}$ denoting the lower and upper bounds of the $(1-\alpha)$ PI \cite{Nowotarski2018RecentForecasting}.
The PICP measures the proportion of actual values that fall within the predicted interval.

Moreover, to evaluate the overall calibration of a forecasting model across a range of confidence levels, we define $MAACE$:
\begin{eqnarray*} %\label{em:maace}
    MAACE = \dfrac{1}{N_{\alpha}}\sum_{\alpha\in \mathcal{A}} |ACE_{1-\alpha}|,
\end{eqnarray*}
where $ACE_{1-\alpha}  = PICP_{1-\alpha} - PINC_{1-\alpha}$ is the average calibration error, and $PINC$ is the prediction interval nominal coverage \cite{Nowotarski2018RecentForecasting}. 
The sharpness of the predicted quantiles is measured using the ${CRPS}$, which we calculate with the pinball loss function \cite{Nowotarski2018RecentForecasting}:
\begin{eqnarray*} %\label{em:crps}
    CRPS_{d,h} &=& \int_{-\infty}^{\infty} (\hat{F}(x)-\mathds{1}_{\{p_{d,h}\leq x\}})^2 dx = \int_0^1 PS_{d,h}^q dq, \\
    &\approx& \dfrac{2}{99} \sum_{q=1}^{99} PS_{d,h}^q
\end{eqnarray*}
with
\begin{eqnarray*}
    PS_{d,h}^q = 
    \begin{cases}
        (1-q)(\hat{p}_{d,h}^q - p_{d,h}), & \text{if } p_{d,h} < \hat{p}^q_{d,h},\\
        q(p_{d,h} - \hat{p}_{d,h}^q), & \text{otherwise}
    \end{cases}
\end{eqnarray*}
where $\hat{p}^q_{d,h}$ denotes the forecast of the $q$th quantile of $p_{d,h}$.
The scaling factor of 2 is commonly omitted in practice \cite{Lipiecki2024PostprocessingRegression}, a convention we follow here.

The MPIW is defined as \cite{khosravi2010prediction}:
\begin{eqnarray*}
    {MPIW} = \dfrac{1}{N_{d,h}}\sum_{i=1}^{N_{d,h}} U_{d,h}^{(1-\alpha)}-L_{d,h}^{(1-\alpha)}.
\end{eqnarray*}

Finally, we compare model accuracies pairwise using the Diebold-Mariano (DM) test \cite{diebold2002comparing}, with CRPS as the loss function.

% Furthermore, we also perform the pair-wise statistical test for the model's accuracy using the Diebold-Mariano (DM) test \cite{diebold2002comparing}. 
For the mathematical formulation of the DM test, please refer to Appendix \ref{appendix:dbTest}.

\subsection{Error Analysis}
\label{ss:error_analysis}
% \textcolor{teal}{
% %Since the table has some space left, we could add the Kupiec test (50\% and 90\% (98\%?)) like in \cite{epf_marcjasz2023distributional}.
% }

We begin by evaluating the point prediction performance of the models. According to Table \ref{tab:4MAE}, the LEAR model achieves the best overall point prediction performance, with an MAE of $12.671$ and an RMSE of $21.414$, outperforming all other models. 
As expected, the naive models yield significantly higher errors compared to the NN-based and statistical approaches. 
Among the neural networks, the DDNN already achieves a considerable reduction in MAE and RMSE relative to the naive models. 
The ensemble models further improve performance, reducing MAE and RMSE compared to the DDNN. 
While increasing the number of models in the ensemble brings only marginal performance improvements, it noticeably reduces the standard deviation of the error metrics. 
MC Dropout models lead to even greater improvements, with reductions in MAE and RMSE of up to $6.4$\% and $4.9$\% relative to the DDNN. 
The number of forward passes in the MC Dropout models shows only a marginal positive effect on performance. 
Notably, the EvDNN achieves the lowest MAE among the neural models, with a $7.4$\% reduction compared to the DDNN.

When examining the interval forecasts, it becomes clear that the EvDNN model struggles to capture uncertainty accurately.  
It produces a relatively high MAACE of about $44$\% and the third-highest CRPS among all models.  
Similar to the point forecasts, the DDNN model significantly improves the probabilistic metrics compared to the naive baselines, and both ensemble and MC Dropout extensions further enhance performance.  
Specifically, they reduce the CRPS by about $4$\% and $7$\%, respectively, compared to the DDNN.  
For MAACE, these methods achieve notable improvements as well.  
The ensemble reduces MAACE by $3$ points (a $49$\% improvement) and MC Dropout by $3.6$ points (a $67$\% improvement), again relative to the DDNN.  
Interestingly, increasing the number of ensemble models has a strong impact on CRPS but little effect on MAACE, while the opposite trend is observed with the number of forward passes in the MC Dropout models.  
Another point worth noting is the relatively high standard deviation of MAACE produced by the DDNN, which is considerably larger than that of the other models.  
The LEAR-GARCH model achieves the best overall forecasting performance in terms of CRPS, with a CRPS of $4.662$, followed closely by the LEAR-CP model at $4.666$ and the LEAR-QRA model at $4.670$.  
When assessing the uncertainty measure MAACE, the top-performing model is the LEAR-CP with a value of $1.305$, followed by the MCD30-CP at $1.735$ and the MCD30 at $1.962$.  
\input{tab-4MAE}

\input{tab-4PICP}
\input{tab-4MPIW}
\begin{figure}[ht!]
% \begin{table}[H]
    \centering
    \includegraphics[width = 8.5cm, height = 6cm]
    {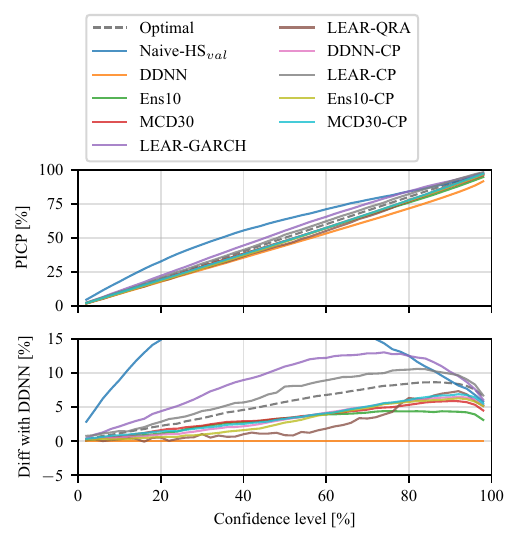}
    \caption[Model evaluation using PICP across different confidence intervals.]{Models are evaluated using PICP across confidence intervals ranging from 2\% to 98\%. The top subplot shows the average results over 10 independent runs, with standard deviations indicated by dotted lines. The bottom subplot displays the difference in performance between the DDNN and the other models.}
    \label{fig:4PICP}
\end{figure}
\begin{figure}[ht!]
% \begin{table}[H]
    \centering
    \includegraphics[width = 8.5cm, height = 6cm]
    {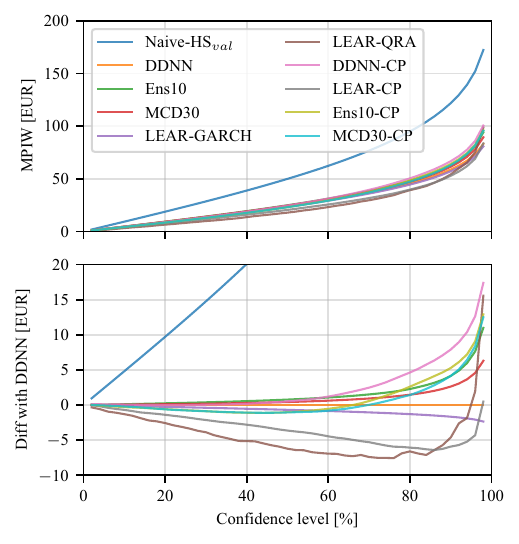}
    \caption[Model evaluation using MPIW across different confidence intervals.]{Models are evaluated using MPIW across confidence intervals ranging from 2\% to 98\%. The top subplot shows the average results over 10 independent runs, with standard deviations indicated by dotted lines. The bottom subplot displays the difference in performance between the DDNN and the other models.}
    \label{fig:4MPIW}
\end{figure}

Next, we examine the PICP metric across different confidence levels to assess how well the predicted intervals are calibrated.
% Next, we examine the PICP metric across different confidence intervals to better understand how varying confidence levels affect the MAACE metric
The corresponding results are presented in Table \ref{tab:4PICP} and Fig. \ref{fig:4PICP}.
It should be noted that, of all the figures, there are 10 models under consideration, whereas the tables in this section include 15 models in total. Five models, namely Naive-HS$_{train}$, Ens5, MCD10, EvDNN, and EvDNN-CP, are not shown in Fig. \ref{fig:4PICP} and Fig. \ref{fig:4MPIW}, since we consider only one better-performing model from each category, and the EvDNN model struggles to capture uncertainty.
The naive models consistently produce PICP values higher than the expected nominal coverage, indicating underconfident intervals. 
The same holds for LEAR-GARCH and LEAR-CP. 
In contrast, all other models tend to produce overconfident PIs. 
Both the ensemble and MC Dropout models show improved PICP values compared to the DDNN, especially as the confidence level increases. 
The greatest improvements are observed around the $90$\% confidence level. 
Up to the $60$\% confidence level, the performance of the ensemble and MC Dropout models is quite similar. 
Moreover, both the ensemble and MC Dropout models significantly reduce the standard deviation of results across independent runs by about half compared to the DDNN, indicating more stable performance. 
After the $60$\% confidence level, the NN-CP models surpass both the ensemble and MC Dropout models. 
At about the $80$\% confidence level, the LEAR-QRA begins to surpass the previous models, while the LEAR-CP achieves the best overall coverage, as reflected in the MAACE metric.

As shown in Table \ref{tab:4MPIW} and Figure \ref{fig:4MPIW}, the MPIW increases almost linearly with the confidence level up to around the $70$\% confidence interval.
As expected, the naive model produces significantly wider PIs on average, consistent with its high PICP values.
In contrast, the other models generate considerably narrower intervals, roughly half the width of those from the naive model.
The exception is the EvDNN model, which produces very sharp but less reliable PIs.
Notably, LEAR-QRA consistently produces the narrowest PIs across all confidence levels among the traditional and NN-based models (excluding the EvDNN), with MPIW values of $17.88$ at $50$\% confidence and about $39.18$ at $80$\%.
At $90$\% and $98$\% confidence, however, LEAR-CP and LEAR-GARCH perform best.
These values are substantially lower than those of DDNN, which produces intervals of $24.10$ and $45.78$ at the $50$\% and $80$\% levels, respectively.
The ensemble with 5 models shows a high standard deviation across runs, suggesting that its performance is sensitive to the specific composition of the ensemble.
Up to the $50$\% confidence level, the MPIW values of the ensemble and MC Dropout models (excluding Ens5), as well as the DDNN, are quite similar. Beyond this point, the ensemble and MC Dropout models begin to diverge from the DDNN.
At the $90$\% confidence level, where the highest gains in PICP are observed, the Ens10 and MC30 models improve PICP by $4.36$ and $5.89$ percentage points, respectively, in exchange for increases in MPIW by $4.14$ and $2.63$, reflecting the trade-off between interval width and coverage.
Between the two, MC Dropout performs better overall.

\newpage
\subsection{Empirical Analysis of Diebold-Mariano Test Results}
% TODO (AD): Like discussed, one phrase that models are neglected

% There are also papers where not the whole CRPS is tested but only the "important" significant values, for example $\alpha\in\{0.01,...,0.1\}\cup\{0.9,...,0.99\}$ (Pinball Score), e.g. \cite{epf_lipiecki2024postprocessing}.

% The empirical p-values are shown as a heatmap in Fig.~\ref{fig:dm_test}, where each entry $p_{ij}$ denotes the p-value from the one-sided DM test comparing model $M_i$ (row model) against model $M_j$ (column model). % same in line 696f
{Now we discuss the results of applying the Diebold-Mariano test ('DM test'), see \cite{diebold2002comparing}} for the test.
In this and in the following section, {we leave out} the following five models {from our analysis}
%for the DM test evaluation and in the trading strategies, namely,
Naive-HS$_{train}$, Ens5, MCD10, EvDNN, and EvDNN-CP, since we only want to consider {well-performing models} from each category. {We choose a significance level of $\alpha = 0.05$. We decide to not do a multiple testing correction (e.g. Bonferroni).}
%and the EvDNN model struggled to capture uncertainty.

\begin{figure}[htb]
    \centering
    \includegraphics{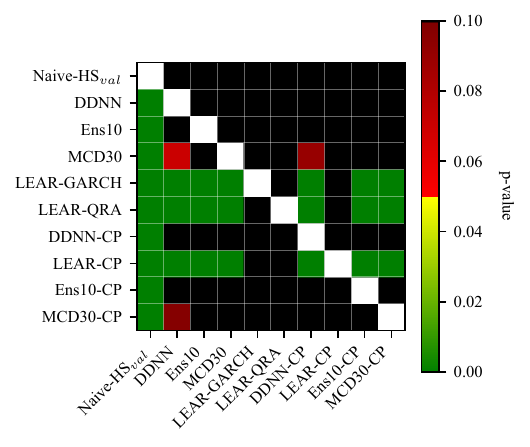}
    \caption{Heatmap of pairwise Diebold–Mariano test $p$-values using CRPS loss. Values below 0.05 mark significant differences in predictive accuracy. Results are averaged over 10 independent runs, yielding 100 comparisons per model pair across runs.}
    \label{fig:dm_test}
\end{figure}

The $p$-values in the DM test show whether the two models under consideration are significantly different in terms of prediction accuracy. 
However, they do not provide any information about models {performance} superiority. To extract this information, one has to analyze the test statistic. 
The mathematical form of the DM test and the table of the test statistic are given in Appendix~\ref{appendix:dbTest} in Table~\ref{tab:dm-matrix}. 
Table~\ref{tab:dm-matrix} reports the corresponding DM statistics \(d_{i,j}\) (based on CRPS loss), where \(d_{i,j}<0\) means \(M_i\) attains lower average loss than \(M_j\) (row is better), \(d_{i,j}>0\) means the opposite (column is better), \(|d_{i,j}|\) quantifies the strength of the advantage, and values numerically near \(0\) indicate practical equivalence. 
In Fig.~\ref{fig:dm_test}, we show the pairwise $p$-values (row model \(M_i\) vs.\ column model \(M_j\)), with significance level $0.05$. 
Entries whose numeric values on the color bar are $< 0.05$ indicate that the observed difference between the two models is statistically significant, whereas larger $p$-values $> 0.05$ indicate no detectable difference.
The diagonal elements are {left blank}
%marked as $\text{nan}$
since models are not tested against themselves. 

Each model is compared with every other model across 10 independent runs.
The results show that the three LEAR–type models form a statistically tied top tier: LEAR–GARCH, LEAR–QRA, and LEAR–CP {performances} are mutually indistinguishable, with near-zero differentials and non-rejections in all pairwise directions.
The pairwise DM statistics given in Table \ref{tab:dm-matrix} in Appendix \ref{appendix:someGraphics} support this claim.
For further clarity on the $p$-values, we also provide their numerical values in Table \ref{tab:dm-pvals} {in the Appendix}.
%(e.g., ${DM}({\text{LEAR-GARCH},\text{LEAR-QRA}})=-0.072$ with $p=0.471$, ${DM}({\text{LEAR-CP},\text{LEAR-GARCH}})=0.049$ with $p=0.520$, and ${DM}({\text{LEAR-CP},\text{LEAR-QRA}})=-0.039$ with $p=0.484$; see Tables~\ref{tab:dm-matrix}–\ref{tab:dm-pvals}).
%$DM(.,.)$ represents the statistic for two models under consideration and is computed as shown in Appendix \ref{appendix:dbTest}.
However, each of these three models dominates every non-LEAR alternative by a wide and statistically {significant} margin.
%for instance, ${DM}({\text{LEAR-CP},\text{Ens10}})=-4.921$ with $p<0.001$, ${DM}({\text{LEAR-GARCH},\text{DDNN}})=-5.763$ with $p=0.000$, and ${DM}({\text{LEAR-QRA},\text{MCD30-CP}})=-4.839$ with $p<0.001$, with analogous results across all such pairs.
All non-LEAR models (DDNN, Ens10, MCD30) and their conformal/MC Dropout variants (DDNN-CP, Ens10-CP, MCD30-CP) are statistically indistinguishable from one another at $\alpha=0.05$; however, directional tendencies do exist {as MCD30 seems to be performing better, generally, although this difference is not significant.}
%(e.g., ${DM}({\text{MCD30},\text{Ens10}})=-1.748$, ${DM}({\text{Ens10},\text{DDNN}})=-2.432$), but the associated $p$–values are too large to claim superiority ($p=0.130$ and $p=0.192$, respectively).
Conformal and MC Dropout models do not yield measurable gains (in terms of CRPS) over their corresponding base models.
%${DM}({\text{MCD30-CP},\text{MCD30}})=0.438$ with $p=0.624$, ${DM}({\text{Ens10-CP},\text{Ens10}})=0.299$ with $p=0.578$, and ${DM}({\text{DDNN-CP},\text{DDNN}})=-0.495$ with $p=0.422$, all failing to reject.
Finally, every method strictly improves upon the naive baseline.
%(e.g., ${DM}({\text{DDNN},\text{Naive-HS}{val}})=-6.273$, ${DM}({\text{Ens10},\text{Naive-HS}{val}})=-6.984$,  ${DM}(\text{LEAR-CP},$ $\text{Naive-HS}_{val}) =-10.235$, all with $p<0.001$).

In summary, (i) there is no single best model within the LEAR family, but together they form a statistically superior performance frontier; (ii) non-LEAR methods are statistically tied with one another and uniformly inferior to the LEAR models; and (iii) naive forecasting {is} decisively {outperformed by} all other approaches.

\subsection{Trading Strategies}
\label{subsec:Trading}

A different angle to evaluate the goodness of fit of a specific forecasting method is to integrate it into a trading strategy and observe the financial success. 
We follow the suggestion of \cite{Marcjasz2018SelectionForecasting} and aim to optimally control a battery storage system based on interval forecasts of the German day-ahead power price. 
Let the battery have a capacity of $B = 2 \, MWh$ and a charging/discharging efficiency of $\xi = 90\%$. 
Each trading day $d$, with $d = 1,\ldots,D$ and $D$ being the number of trading days, we pick an hour $h_{buy}$ for buying $1\, MWh$ and another hour $h_{sell}$ for selling it. 
We start with an initial charge of $B_{d} = 1 \, MWh$ on day $d=1$. 
Transaction costs are ignored.

We consider a quantile-based strategy that uses PIs generated by each of the methods tested in Section \ref{ss:error_analysis}, as well as the perfect foresight strategy, where we know the true prices, as the profit's upper boundary. 
The quantile-based strategy works by setting limits, which are given by the PIs: the selling price is equal to the lower boundary of the highest price's PI $L_{d,h_{sell}}^{(1-\alpha)}$; correspondingly, we obtain the purchasing price via the upper boundary of the lowest forecasted price's PI $U_{d,h_{buy}}^{(1-\alpha)}$ given a $1-\alpha$ confidence level. We perform both trades if and only if
    \begin{eqnarray} \label{eq:limit_profitable}
        \xi\, L_{d,h_{sell}}^{(1-\alpha)} - \dfrac{1}{\xi} \, U_{d,h_{buy}}^{(1-\alpha)} > 0,\quad h_{sell}\neq h_{buy},
    \end{eqnarray} 
with $h_{sell}$/$h_{buy}$ being the hour of the highest/lowest predicted price. 
Since we use limits, trading is not mandatory, and we might end a day with $B_d=2\, MWh$ or $B_d=0\, MWh$.
In that situation, to avoid staying at the battery's margins, we place two limited and one unlimited bid the next day by solving the following optimization problem:
\begin{itemize}
        \item[(a)] If $B_d=2MWh$: Solve
    \begin{eqnarray*}
        \max_{h_{sell,1},h_{sell,2},h_{buy}} \xi\, \hat{p}_{d,h_{sell,1}} + \xi\, \hat{p}_{d,h_{sell,2}} - \dfrac{1}{\xi} \, \hat{p}_{d,h_{buy}},\\
        h_{sell,1}\neq h_{sell,2}\neq h_{buy}, \quad h_{sell,1}<h_{buy}.
    \end{eqnarray*}
    The trades are executed if:
    \begin{eqnarray} \label{eq:limit_profitable_B2}
        \xi\, \hat{p}_{d,h_{sell,1}} + \xi\, L_{d,h_{sell,2}}^{(1-\alpha)} - \dfrac{1}{\xi} \, U_{d,h_{buy}}^{(1-\alpha)}>\xi\, \hat{p}_{d,h_{sell,3}}, 
    \end{eqnarray}
    where $h_{sell,3}$ is the hour with the highest predicted price. If this condition is not met, an unlimited sell order is placed at $h_{sell,3}$.
    \item[(b)] If $B_d=0MWh$: Solve
    \begin{eqnarray*}
        \max_{h_{sell},h_{buy,1},h_{buy,2}} -\dfrac{1}{\xi}\, \hat{p}_{d,h_{buy,1}} + \xi\, \hat{p}_{d,h_{sell}} - \dfrac{1}{\xi} \, \hat{p}_{d,h_{buy,2}},\\ 
        h_{sell}\neq h_{buy,1}\neq h_{buy,2}, \quad h_{buy,1}<h_{sell}.
    \end{eqnarray*} 
    The trades are executed if:
    \begin{eqnarray} \label{eq:limit_profitable_B0}
        -\dfrac{1}{\xi}\, \hat{p}_{d,h_{buy,1}} + \xi\, L_{d,h_{sell}}^{(1-\alpha)} - \dfrac{1}{\xi} \, U_{d,h_{buy,2}}^{(1-\alpha)}>-\dfrac{1}{\xi}\, \hat{p}_{d,h_{buy,3}},
    \end{eqnarray}
    where $h_{buy,3}$ is the hour with the lowest predicted price. If this condition is not met, an unlimited buy order is placed at $h_{buy,3}$.
\end{itemize}

%\textcolor{blue}{
%\begin{itemize}
%    \item Due to the use of limit orders, trades may not be executed. The sell order is only accepted if the limit price is below the actual price  and the buy order is only accepted if the limit price is above the actual price ($U_{d,h_{buy}}^{(1-\alpha)}>p_{d,h_{sell}}$). 
%    If either the sell or buy order is not executed, the battery ends the day either with $B_d=2MWh$ or $B_d=0MWh$. In that case, we place two limited orders and one unlimited order on the following day. However, we cannot simply choose the hours with the lowest and highest predicted prices. Instead, we need to solve a linear optimization problem to determine the optimal trading hours that maximize expected profit:
    
    %\item Unlimited bids: This strategy places a buy and a sell order every day based only on predicted prices, without price limits. Orders are always executed at the actual Day-Ahead price. The hours $h_{sell}$ and $h_{buy}$ are selected based on the lowest and highest predicted prices. A trade is made only if it's expected to be profitable:
    %\begin{equation}
    %    \xi\, \hat{p}_{d,h_{sell}} - \dfrac{1}{\xi} \, \hat{p}_{d,h_{buy}}>0,\quad h_{sell}\neq h_{buy}.
    %\end{equation}
 %   \item Fixed hours: This strategy chooses the same trading hours each day, based on the average lowest and highest price hours over a historical reference period. It does not use forecasts for decision-making.
 %   \item Optimal bids: This theoretical benchmark assumes perfect forecasts and selects the actual best hours to buy and sell each day based on observed prices.
%\end{itemize}

%}
To make sure we end at $B_d=1$, we trade unilaterally if $B_d=0$ or $B_d=2$ on the last day of the trading period. 
Eventually, we evaluate performance based on the total profit over the test period and the per-transaction profit, which is calculated by dividing the total profit by the number of executed trades.

\begin{figure}[htbp]
    \centering
    \includegraphics%[width = 8.5cm, height = 5cm]
    {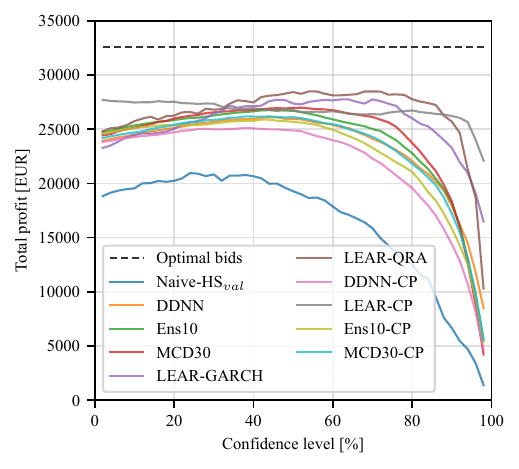}
      \caption[Total profit of models across different confidence intervals.]{Total profit of models is shown for confidence intervals ranging from 2\% to 98\% using the quantile-based strategy. For comparison, the optimal bids strategy is also included. Results are averaged over 10 independent runs.}
      % TODO (AL): check all captions, since we left out a few strategies.
    \label{fig:4TProfit}
\end{figure}

\begin{figure}[htbp]
% \begin{table}[H]
   \centering
   \includegraphics
   {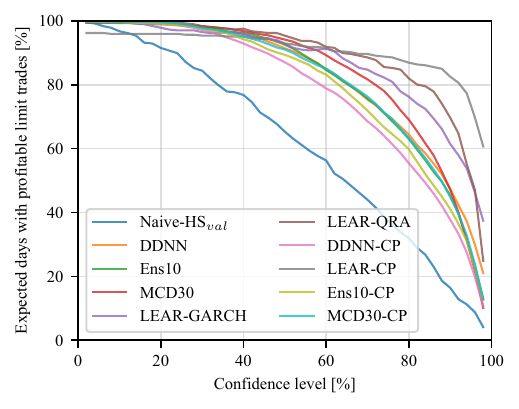}
   \caption[Expected number of days with profitable limit trades across different confidence intervals.]{The expected number of days with profitable limit trades is shown for each model across confidence intervals ranging from 2\% to 98\% using the quantile-based strategy. Results are averaged over 10 independent runs.}
   \label{fig:4expd}
\end{figure}

In Fig. \ref{fig:4TProfit}, the total profits for all tested strategies are shown depending on the chosen confidence level. 
We see that each model's total payoff peaks at a certain confidence level, after which it declines. 
Note that this peak confidence level differs by model. 
The NN-based model's profit, for example, culminates around $50$\%. 
For the naive model, the peak is around $30$\%, and around $85$\% and $90$\% for the LEAR-QRA and LEAR-GARCH models, respectively. 
The decline is caused by the widening PIs, which lead to fewer trading days identified as profitable, resulting in fewer trades, as shown in Fig. \ref{fig:4expd}.
Both the ensemble and MC Dropout models outperform the DDNN model in terms of total profit at the $90$\% confidence level, despite having wider PIs as shown in Fig. \ref{fig:4MPIW}.
At the $70$\% confidence level, where the difference in total profit is most significant, the ensemble model achieves a profit that is $3.7$\% higher, while the MC Dropout model achieves an $8.4$\% higher profit compared to the DDNN model.

Notably, the LEAR-QRA model outperforms all other models between confidence levels of approximately $35$\% and $85$\%, until the PIs become too wide.
Interestingly, below $35$\% and again between $85$\% and $95$\%, the LEAR-CP model achieves the highest total profits.
%% old
% The expected number of days with profitable limit trades for EvDNN and LEAR-GARCH remains high at high confidence intervals. 
% As a result, there is almost no decline in total profit. 
% This results in both models achieving the highest total profit, starting around $90$\%.
% However, it is important to note that the PIs are significantly less reliable.

\begin{figure}[htbp]
% \begin{table}[H]
    \centering
    \includegraphics{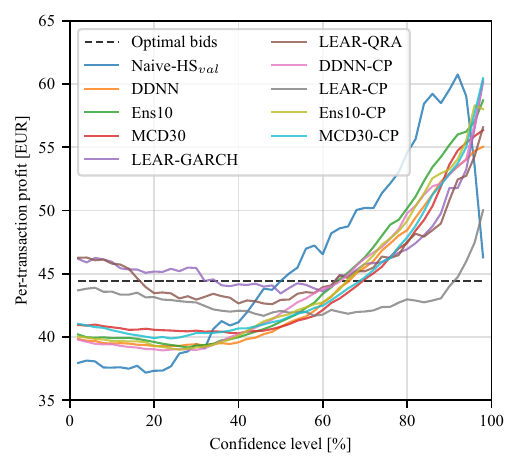}
    \caption[Per-transaction profit of models across different confidence intervals.]{Per-transaction profit of models is shown for confidence intervals ranging from 2\% to 98\% using the quantile-based strategy. For comparison, the optimal bids strategy is also included. Results are averaged over 10 independent runs.}
    \label{fig:4ptp}
\end{figure}

Fig. \ref{fig:4ptp} shows the profit per transaction. 
The per-transaction profit in quantile-based strategies tends to increase with higher confidence levels.
In contrast, the statistical models show an initial decline up to around the $50$\% confidence interval, after which their profits begin to rise.
Interestingly, the LEAR-CP model remains relatively stable up to around $90$\%, before it also starts to increase.
After reaching the $50$\% confidence level, the naive model begins to outperform the others in terms of per-transaction profit. 
This suggests that although it executes fewer trades, the trades it does make are generally more profitable. 
However, its profit curve starts to decline at the $90$\% confidence level. 
% Until approximately the $40$\% confidence level, all NN models demonstrate similar per-transaction profits. 
% Beyond this point, their performance begins to diverge. 
% Notably, the DDNN model initially positions itself between the ensemble and MC Dropout models. 
% However, the MC Dropout model surpasses the DDNN model at around the $90$\% confidence level. 
Notably, the NN models and their CP counterparts exhibit very similar per-transaction profits.
The statistical models demonstrate higher per-transaction profits than all other models during the first half of the confidence level. 
Afterward, LEAR-GARCH and LEAR-QRA tend toward the performance of the other models, while LEAR-CP remains significantly lower than the rest.
% The LEAR-QRA consistently achieves greater per-transaction profit across all confidence levels compared to the optimal bids strategy. 
% Its per-transaction profit remains steady until the $80$\% confidence interval, after which it begins to increase, surpassing all other models around the $95$\% confidence interval. 
% In contrast, the LEAR-GARCH model experiences a slight decline before it starts to rise again at around the $95$\% confidence interval.

The tables for Fig.~\ref{fig:4TProfit}, Fig.~\ref{fig:4expd}, and Fig.~\ref{fig:4ptp}, along with their corresponding plots against PICP, can be found in Appendix~\ref{appendix:someGraphics}.

\section{Discussion and  Critical Reflection}
\label{sec:discussion}

In general, it shows that there is no superior method that performs best across all error measures. 
However, there is a certain tendency towards LEAR-based models, which show surprisingly good performance in the presence of state-of-the-art machine learning benchmarks, despite being much simpler and requiring considerably less computational power. 
This indicates, at least given the setup of this case study, that the additional complexity of machine learning models is not paying off. The case study shows that the LEAR-CP model has, on average, the most reliable PIs across all models, with a MAACE of $1.305$, whereas the best machine learning model, MC Dropout, achieves a MAACE of $1.735$. 
At a confidence level of around $85\%$ and higher, the LEAR-QRA starts to outperform the MD model, achieving a better PICP. 
%The LEAR-QRA is the best statistical model, with a MAACE of 4.766, but the MAACE is higher than all ensemble and Monte Carlo dropout models.
The evidential neural network failed to capture the uncertainty. 
One possible reason could be the zero-gradient problem for evidential regression models; see \cite{EvDL_ye2024uncertainty} for more information.
The LEAR-GARCH also had difficulties in providing reliable PIs, as did the naive approach. The problem is that their uncertainty predictions solely depend on a historical price period. {The superior performance of the LEAR-based models may be explained by the fact that they are refitted for every prediction. This is only possible because of the relatively low computational cost.}

We notice that regarding the point forecasts, the statistical model LEAR performs better than all other models. 
A possible reason could be the use of the rolling window scheme in contrast to the NN models, where the training data is fixed.
%\textcolor{red}{But it would be massively time consuming to introduce a rolling window for ML models, right? So actually, that's not really realistic, or is it? I mean, a rolling window is used to some extent as normally the previous few days are fed to the trained network, right?}
%\textcolor{teal}{Yes, but normally you have the time to compute the model for the next day..}

Our study indicates that quantifying both data and model uncertainty, particularly for deep learning models, pays off. 
Ensemble-based methods help to reduce MAE by $0.63$ ($-3.67$\%) for Ens10 and $1.09$ ($-6.35$\%) for MCD30, and RMSE by $0.74$ ($-2.89$\%) for Ens10 and $1.26$ ($-4.90$\%) for MCD30 compared to DDNN. 
The run-to-run variability of errors is also smaller. 
For MAE, the standard deviation drops from $0.767$ (DDNN) to $0.211$ (Ens10, $-72$\%) and $0.412$ (MCD30, $-46$\%). 
MC Dropout helps reduce both MAE and RMSE compared to DDNN. Interestingly, evidential deep neural networks show the lowest MAE but generate less reliable PIs compared to their benchmarks.

We find that improvements in reliability come at the cost of wider PIs, i.e., reduced sharpness. 
The naive model produced underconfident, i.e., too wide PIs, whereas NN-based models tend to be overconfident. 
However, using ensembles and MC dropout considerably improves PICP values, particularly at high confidence levels, and reduces PICP variability across runs.

Interestingly, incorporating model uncertainty into DDNN models also helps improve their point forecasts. 
This is due to the procedures to quantify and estimate model uncertainty, and not due to modeling uncertainty itself.

In terms of economic benefits, good performance in interval forecasting is no guarantee of actual profit in daily trading. The reason is that we are penalizing the profits because of the uncertainty that comes with them. 
Uncertainty is included in the objective function in terms of variance, and it is considered to be bad. 
But in fact, volatility brings profits. 
We see that the quantile-based strategy generated fewer trades as the confidence level increased, since wider predicted limits reduce expected profitable opportunities. 
Wider, less precise intervals typically reduce the number of trades but increase per-transaction profit.

Eventually, there are a few limitations of the tested models that need to be mentioned. For example, we see that it pays off to use ensemble models for generating PIs. 
However, the choice of the required hyperparameters is rather tricky. Ensemble forecasting requires specifying the number of ensemble members, which is difficult to determine. 
Additionally, the composition of these members is significant. 
Weak members can harm overall performance, and equal-weight aggregation might be suboptimal.

In the statistical models, we implemented a rolling window approach for calibrating the parameters. 
Although this is beneficial in many situations, it causes problems during structural breaks. 
Moreover, the window size needs to be chosen beforehand, and this choice is always somewhat arbitrary. 
Statistical models are lighter and interpretable but can be brittle under regime shifts and heavy tails if distributional assumptions are too restrictive. 
ML models are adaptive but need careful regularization and calibration to avoid over- or underconfidence. 
Rather than choosing one, the evidence suggests a hybrid path: use LEAR-type models for baseline calibration and coverage control, and combine them with deep learning models when diagnostic plots indicate systematic misspecification.

In this study, Gaussian innovations were used for comparability. Moreover, a natural extension is to test heavy-tailed and skewed families (Student-$t$, skew-$t$, generalized error, generalized hyperbolic or {JSU}) within LEAR-GARCH. 
This could retain the calibration advantages while improving tail coverage, again providing a way to strengthen statistical models without diminishing the role of ML.

The tables indicate that carefully specified statistical models, particularly LEAR-CP, LEAR-QRA, and LEAR-GARCH, deliver reliable and sharp uncertainty quantification across the nominal levels we care about. 
The DM test results also support this claim. 
For example, LEAR-CP attains the best MAACE ($1.305$) and competitive coverage at $90$-$98$\% (e.g., PICP$_{98\%}=98.18\%$ with MPIW$_{98\%}=83.60$), while LEAR-QRA yields the narrowest mid-range intervals (best MPIW at $50$-$80$\%) with solid coverage. 
These are attractive operating points for system operators who value calibration and interpretability. 
Importantly, these gains do not imply that ML is "bad"; rather, they show that a well-regularized linear-quantile/expectile framework with autoregressive structure can already capture a large share of predictable variation in prices.

ML approaches clearly add value: the ensemble and MC dropout approaches improve DDNN’s point and probabilistic metrics (e.g., CRPS reduction of $4.8$–$7.7$\% vs. DDNN) and markedly stabilize training outcomes (large reductions in MAE standard deviation). 
In parallel, LEAR-based models set a strong baseline with transparent coefficients, fast retraining in rolling windows, and consistent coverage. 
A pragmatic takeaway is to treat LEAR-type models as robust, low-latency baselines and to prioritize ML-based uncertainty procedures when nonlinearity or feature interactions matter.

Electricity prices exhibit strong autoregressive structure, intraday seasonality, and heteroskedasticity. 
Linear quantile/expectile regressions with lagged terms and simple volatility updates exploit exactly these features; when the signal-to-noise ratio is moderate and exogenous effects are mostly linear in impact, the marginal benefit of deep nonlinearity can be limited. 
This does not negate the usefulness of ML; it clarifies when ML’s flexibility is most likely to pay off, e.g., with richer exogenous stacks, structural breaks, or nonlinear regime interactions.

For risk-aware bidding, well-calibrated statistical intervals provide immediate knobs for setting quantile-based limits that translate to risk-adjusted expected profit. 
ML methods remain valuable for discovering nonlinear predictive features that can then be distilled into simpler statistical decision rules (e.g., via quantile targets). 
This "learn nonlinearity, decide linearly" pipeline can balance robustness, transparency, and performance.

\section{Conclusion}
\label{sec:conclusion}

In this article, we test a wide range of methods on their capability to produce reliable interval forecasts for hourly power prices. 
In this context, we analyze whether forecasting accuracy improves if model and data uncertainty are explicitly considered using an adequate model, such as a Bayesian approach. 
An extensive model comparison based on various goodness-of-fit measures is performed, whereby both statistical and machine learning models are tested. 
It shows that there is no model that performs best across all measures. Moreover, having also considered economic implications in the context of a trading scheme, we see that increasing the precision of interval forecasts does not necessarily increase profits. 
With regard to PIs, it shows that it pays off to consider uncertainty, whereby we see that a combined focus on both aleatoric and epistemic uncertainty is required. 
Ensemble models help to reduce both MAE and RMSE. 
However, they are outperformed by statistical LEAR models, where a combination with GARCH volatility proves to be fairly reliable. 
From a statistical standpoint, the strength of the LEAR family is calibration. 
Their interval systems are directly tied to quantile/expectile targets, which often translates into well-behaved Probability Integral Transform (PIT) diagnostics and reliable Coverage vs. Nominal (CvN) curves.
A constructive next step, without undermining ML, would be to use LEAR-based forecast distributions as calibration anchors for neural outputs (e.g., post-hoc quantile mapping or conformalized residuals). Eventually, we are aware of a few limitations of our study and paths for future research. First, only the German power market is tested, and it remains to be verified how the models perform on other markets. 
With regard to machine learning models, we could test hybrid uncertainty quantification methods, i.e., we could try to combine complementary methods in order to capture both epistemic and aleatoric uncertainty. 
Post-hoc calibration is not yet fully explored \cite{Song2019DistributionRegression}. 
Other methods, such as adaptive ensemble sizing and performance-based ensemble member selection, might also help to improve PIs. 
Furthermore, more complex distribution families, such as the generalized hyperbolic distribution for modeling randomness, could be explored, or alternatively, distribution-free approaches. 
Eventually, considering the comparably poor financial performance of the tested complex models, one needs to explore new possibilities that arise from the increasing availability of storage capacity on the German market. 
Here, uncertainty quantification could be more beneficial than in the currently tested scenario.

\section*{Acknowledgment}
Part of this work is derived from the first author’s M.Sc. thesis \cite{lededev_thesis} and has been substantially extended and revised for this article.
The authors acknowledge support by the state of Baden-Württemberg through bwHPC. The third author gratefully acknowledges that this research was (partially) funded in the course of TRR 391 Spatio-temporal Statistics for the Transition of Energy and Transport (520388526) by the Deutsche Forschungsgemeinschaft (DFG, German Research Foundation).

\bibliographystyle{IEEEtran}
\bibliography{sn-bibliography}

\appendix
\section{}
\subsection{The Diebold-Mariano Test}
\label{appendix:dbTest}
The Diebold-Mariano (DM) test is designed as follows: Let $e_{1t}$ and $e_{2t}$ denote the forecast errors of Model 1 and Model 2, respectively, at time $t$. 
To quantify the loss associated with these errors, a loss function $g(\cdot)$ is employed. 
Common choices for $g(\cdot)$ include the squared error loss, $g(e_t) = e_t^2$, or the absolute error loss, $g(e_t) = |e_t|$. The differential loss at time $t$ is then defined as $d_t = g(e_{1t}) - g(e_{2t})$.
The null hypothesis ($H_0$) of the DM test posits that the two forecasts exhibit identical accuracy, implying that the expected value of the loss differential is zero:
$$H_0: E[d_t] = 0$$
The alternative hypothesis ($H_1$) asserts that the forecasting accuracies are distinct:
$$H_1: E[d_t] \neq 0$$
The DM test statistic is constructed as:
$$DM = \frac{\bar{d}}{\sqrt{\widehat{\text{Var}}(\bar{d})}}$$
where $\bar{d} = \frac{1}{T}\sum_{t=1}^T d_t$ represents the sample mean of the loss differential over $T$ observations, and $\widehat{\text{Var}}(\bar{d})$ is a consistent estimator of the variance of $\bar{d}$. 
This variance estimator is crucial, as it accounts for potential serial correlation and heteroskedasticity inherent in the loss differential series $d_t$. Under the null hypothesis, the DM test statistic asymptotically follows a standard normal distribution, i.e., $DM \sim N(0,1)$. 
Effective application of the DM test for forecast accuracy assessment necessitates several key prerequisites. 
First, accurate computation of forecast errors for each model is required, typically defined as the difference between the observed value and the forecasted value. 
Second, a clearly defined and appropriate loss function must be specified to quantify the cost of forecasting errors in a manner relevant to the specific application, as the selection of the loss function directly impacts the interpretation of "accuracy." 
Third, as an asymptotic test, the DM test relies on a sufficiently large sample size ($T$) to ensure that the DM test statistic converges to its asymptotic normal distribution, thereby validating the statistical inference. 
Finally, while the original formulation assumed covariance stationarity of the loss differential series ($d_t$), modern extensions of the DM test can accommodate non-stationary series, enhancing its applicability to a broader range of time series data. 
The $p$-value derived from the DM test statistic serves as the basis for inference. 
A $p$-value below a pre-specified significance level ($\alpha$, commonly 0.05) leads to the rejection of $H_0$, indicating a statistically significant difference in forecasting accuracy between the two models.

% \begin{table*}[ht!]
% \tiny
% % \begin{table}[H]
%     \centering
% \label{tab:dm-matrix}
% \setlength{\tabcolsep}{3pt}
% \begin{tabular}{lrrrrrrrrrrrrrr}
% \toprule

% \newpage

\subsection{Supplementary Graphics and Tables}
\label{appendix:someGraphics}

\begin{figure}[htbp]
% \begin{table}[H]
    \centering
    \includegraphics%[width = 8.5cm, height = 5cm]
    {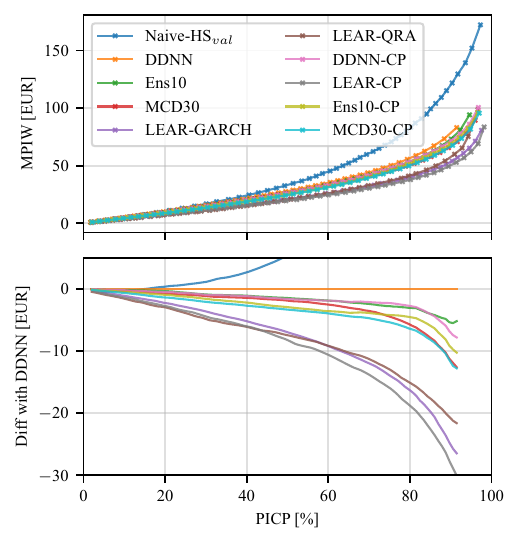}
    \caption[Model evaluation using MPIW across confidence intervals, plotted against PICP.]{Models are evaluated using MPIW across confidence intervals ranging from 2\% to 98\%, and the results are plotted against PICP. The top subplot shows the average results over 10 independent runs, with standard deviations indicated by dotted lines. The bottom subplot displays the difference in performance between the DDNN and the other models, calculated using piecewise linear interpolation of the curves.}
    \label{fig:PICPagainstMPIW}
\end{figure}

\begin{figure}[htbp]
% \begin{table}[H]
   \centering
   \includegraphics%[width = 8.5cm, height = 5cm]
   {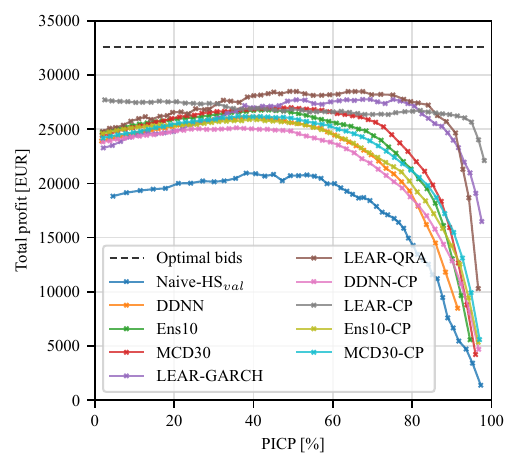}
   \caption[Total profit of models across different confidence intervals, plotted against PICP.]{Total profit of models is shown for confidence intervals ranging from 2\% to 98\% using the quantile-based strategy, and the results are plotted against PICP. For comparison, the optimal bids strategy is also included. Results are averaged over 10 independent runs.}
   \label{fig:4picp_TProfit}
\end{figure}

\begin{figure}[htbp]
% \begin{table}[H]
   \centering
   \includegraphics%[width = 8.5cm, height = 5cm]
   {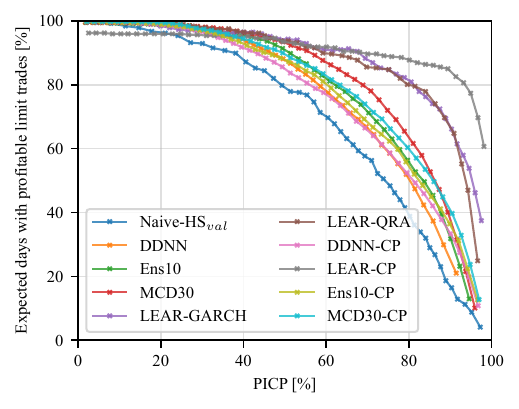}
   \caption[Expected number of days with profitable limit trades across different confidence intervals, plotted against PICP.]{The expected number of days with profitable limit trades (satisfying Equation \ref{eq:limit_profitable}, \ref{eq:limit_profitable_B2}, or \ref{eq:limit_profitable_B0}) is shown for each model across confidence intervals ranging from 2\% to 98\% using the quantile-based strategy and the results are plotted against PICP. Results are averaged over 10 independent runs.}
   \label{fig:4picp_expd}
\end{figure}

\begin{figure}[htbp]
% \begin{table}[H]
   \centering
   \includegraphics%[width = 8.5cm, height = 5cm]
   {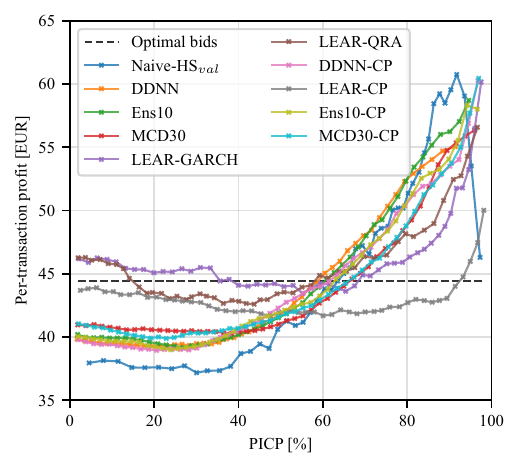}
   \caption[Per-transaction profit of models across different confidence intervals, plotted against PICP.]{Per-transaction profit of models is shown for confidence intervals ranging from 2\% to 98\% using the quantile-based strategy, and the results are plotted against PICP. For comparison, the optimal bids strategy is also included. Results are averaged over 10 independent runs.}
   \label{fig:4picp_ptp}
\end{figure}

This appendix provides additional results for Section~\ref{sec:CaseStudy}.
Figures~\ref{fig:PICPagainstMPIW}–\ref{fig:4picp_ptp} show MPIW, total profit, the number of profitable days, and per-transaction profit plotted against PICP.
Tables~\ref{tab:hp_opt}-\ref{tab:dm-pvals} report optimal hyperparameters, total profit, the number of profitable days, per-transaction profit, DM statistics, and DM $p$-values.

\input{tab-hp_opt}

\input{tab-TProfit}
\input{tab-4exdp}
\input{tab-4ptp}
\input{tab-dbvalues}

\input{tab-p-values}

\end{document}

%% file: tab-4MAE.tex
\newcommand{\meanStdh}[2]{%
  #1{\color{gray}\footnotesize$\pm#2$}%
}

\begin{table*}[ht!]
% \begin{table}[H]
\caption[Model evaluation using point and probabilistic metrics.]{Models are evaluated using point and probabilistic metrics. Results are averaged over 10 independent runs, with standard deviations shown in light gray. The best-performing results are highlighted in bold.}
    \centering
\begin{tabular}{lrrrr}
\toprule
                    % & \glsxtrshort{MAE}                             & \glsxtrshort{RMSE}                            & \glsxtrshort{CRPS}$_{1:99:1}$                & \glsxtrshort{MAACE}$_{2:98:2}$ \\
                    & MAE                             & RMSE                            & CRPS$_{1:99:1}$                & MAACE$_{2:98:2}$ \\
   \midrule
    Naive-HS\(_{train}\) & \meanStdh{24.878}{0.000} & \meanStdh{36.685}{0.000} & \meanStdh{9.823}{0.000}  & \meanStdh{12.842}{0.000} \\
    Naive-HS\(_{val}\)   & \meanStdh{24.878}{0.000} & \meanStdh{36.685}{0.000} & \meanStdh{9.634}{0.000}  & \meanStdh{9.107}{0.000}  \\
    DDNN                & \meanStdh{17.209}{0.767} & \meanStdh{25.684}{0.913} & \meanStdh{6.392}{0.286}  & \meanStdh{5.286}{2.735}  \\
    Ens5                & \meanStdh{16.692}{0.464} & \meanStdh{25.193}{0.685} & \meanStdh{6.276}{0.501}  & \meanStdh{2.398}{1.080}  \\
    Ens10               & \meanStdh{16.577}{0.211} & \meanStdh{24.942}{0.262} & \meanStdh{6.087}{0.069}  & \meanStdh{2.370}{1.043}  \\
    MCD10               & \meanStdh{16.193}{0.427} & \meanStdh{24.549}{0.497} & \meanStdh{5.918}{0.163}  & \meanStdh{2.276}{1.455}  \\
    MCD30               & \meanStdh{16.116}{0.412} & \meanStdh{24.426}{0.447} & \meanStdh{5.897}{0.149}  & \meanStdh{1.962}{1.379}  \\
    EvDNN               & \meanStdh{15.940}{0.697} & \meanStdh{24.566}{0.674} & \meanStdh{7.510}{0.311}  & \meanStdh{43.937}{0.730} \\
    LEAR-GARCH               & \meanStdh{\textbf{12.671}}{0.000} & \meanStdh{\textbf{21.414}}{0.000} & \meanStdh{\textbf{4.662}}{0.000}  & \meanStdh{3.317}{0.000} \\
    LEAR-QRA               & \meanStdh{\textbf{12.671}}{0.000} & \meanStdh{\textbf{21.414}}{0.000} & \meanStdh{4.670}{0.000}  & \meanStdh{2.865}{0.000} \\
    DDNN-CP               & \meanStdh{17.209}{0.767} & \meanStdh{25.684}{0.913} & \meanStdh{6.330}{0.271}  & \meanStdh{1.948}{0.361} \\
    LEAR-CP               & \meanStdh{\textbf{12.671}}{0.000} & \meanStdh{\textbf{21.414}}{0.000} & \meanStdh{4.666}{0.000}  & \meanStdh{\textbf{1.305}}{0.000} \\
    Ens10-CP               & \meanStdh{16.577}{0.211} & \meanStdh{24.942}{0.262} & \meanStdh{6.114}{0.078}  & \meanStdh{2.272}{0.126}  \\
    MCD30-CP               & \meanStdh{16.116}{0.412} & \meanStdh{24.426}{0.447} & \meanStdh{5.942}{0.141}  & \meanStdh{1.736}{0.568}  \\
    EvDNN-CP               & \meanStdh{15.940}{0.697} & \meanStdh{24.566}{0.674} & \meanStdh{5.905}{0.240}  & \meanStdh{2.066}{0.215} \\
\bottomrule
\end{tabular}

\label{tab:4MAE}
\end{table*}

%% file: tab-4PICP.tex
\newcommand{\meanStd}[2]{%
  \shortstack[r]{#1 \\ \color{gray}{\footnotesize{$\pm$#2}}}%
}
\newcommand{\modelName}[1]{%
  \shortstack[r]{#1 \\ \color{white}{\footnotesize{0}}}%
}
\begin{table*}[ht!]
% \begin{table}[H]
\caption[Model evaluation using PICP at different confidence intervals.]{Models are evaluated using PICP at confidence intervals of 50\%, 60\%, 70\%, 80\%, 90\% and 98\%. Results are averaged over 10 independent runs, with standard deviations shown in light gray.The best-performing results are highlighted in bold.}
    \centering
    \begin{tabular}{lrrrrrr}
    \toprule
    % $[\%]$ & \glsxtrshort{PICP}$_{50\%}$ & \glsxtrshort{PICP}$_{60\%}$ & \glsxtrshort{PICP}$_{70\%}$ & \glsxtrshort{PICP}$_{80\%}$ & \glsxtrshort{PICP}$_{90\%}$ & \glsxtrshort{PICP}$_{98\%}$ \\
    $[\%]$ & PICP$_{50\%}$ & PICP$_{60\%}$ & PICP$_{70\%}$ & PICP$_{80\%}$ & PICP$_{90\%}$ & PICP$_{98\%}$ \\
    \midrule
    \modelName{Naive-HS$_{train}$} & \meanStd{67.93}{0.00} & \meanStd{75.20}{0.00} & \meanStd{81.88}{0.00} & \meanStd{87.65}{0.00} & \meanStd{93.89}{0.00} & \meanStd{98.82}{0.00} \\
    \modelName{Naive-HS$_{val}$}   & \meanStd{63.55}{0.00} & \meanStd{71.00}{0.00} & \meanStd{77.87}{0.00} & \meanStd{84.10}{0.00} & \meanStd{\textbf{90.36}}{0.00} & \meanStd{97.29}{0.00} \\
    \modelName{DDNN}               & \meanStd{44.24}{3.51} & \meanStd{53.21}{3.83} & \meanStd{62.28}{3.91} & \meanStd{71.54}{4.02} & \meanStd{81.47}{3.57} & \meanStd{91.47}{2.37} \\
    \modelName{Ens5}               & \meanStd{\textbf{47.67}}{1.50} & \meanStd{57.09}{1.48} & \meanStd{66.58}{1.55} & \meanStd{75.89}{1.41} & \meanStd{85.60}{1.16} & \meanStd{94.35}{0.53} \\
    \modelName{Ens10}              & \meanStd{47.63}{1.37} & \meanStd{57.13}{1.45} & \meanStd{66.48}{1.58} & \meanStd{75.93}{1.57} & \meanStd{85.83}{1.33} & \meanStd{94.60}{0.81} \\
    \modelName{MCD10}              & \meanStd{47.34}{1.91} & \meanStd{56.85}{2.16} & \meanStd{66.49}{2.28} & \meanStd{76.47}{2.21} & \meanStd{86.90}{1.88} & \meanStd{95.71}{0.88} \\
    \modelName{MCD30}              & \meanStd{47.63}{1.65} & \meanStd{\textbf{57.27}}{1.96} & \meanStd{66.92}{2.16} & \meanStd{76.89}{2.02} & \meanStd{87.36}{1.86} & \meanStd{95.98}{0.85} \\
    \modelName{EvDNN}              & \meanStd{4.84}{0.53}  & \meanStd{6.13}{0.76}  & \meanStd{7.63}{0.97}  & \meanStd{9.82}{1.23}  & \meanStd{13.42}{1.72} & \meanStd{22.17}{2.81} \\
    \modelName{LEAR-GARCH} & \meanStd{55.21}{0.00} & \meanStd{65.44}{0.00} & \meanStd{75.18}{0.00} & \meanStd{84.07}{0.00} & \meanStd{91.67}{0.00} & \meanStd{97.55}{0.00} \\
    \modelName{LEAR-QRA} & \meanStd{45.13}{0.00} & \meanStd{55.05}{0.00} & \meanStd{65.61}{0.00} & \meanStd{77.89}{0.00} & \meanStd{88.63}{0.00} & \meanStd{96.65}{0.00} \\
    \modelName{DDNN-CP} & \meanStd{47.43}{0.48} & \meanStd{57.39}{0.51} & \meanStd{67.32}{0.48} & \meanStd{77.48}{0.47} & \meanStd{87.90}{0.36} & \meanStd{96.78}{0.13} \\
    \modelName{LEAR-CP} & \meanStd{52.27}{0.00} & \meanStd{62.15}{0.00} & \meanStd{72.18}{0.00} & \meanStd{\textbf{82.06}}{0.00} & \meanStd{91.38}{0.00} & \meanStd{\textbf{98.18}}{0.00} \\
    \modelName{Ens10-CP} & \meanStd{46.98}{0.18} & \meanStd{57.15}{0.19} & \meanStd{67.17}{0.21} & \meanStd{77.31}{0.25} & \meanStd{87.66}{0.14} & \meanStd{96.66}{0.08} \\
    \modelName{MCD30-CP} & \meanStd{47.53}{0.71} & \meanStd{57.31}{0.98} & \meanStd{\textbf{67.39}}{0.91} & \meanStd{77.59}{0.84} & \meanStd{88.21}{0.57} & \meanStd{96.99}{0.15} \\
    \modelName{EvDNN-CP} & \meanStd{47.21}{0.40} & \meanStd{57.02}{0.41} & \meanStd{67.01}{0.39} & \meanStd{77.18}{0.36} & \meanStd{87.65}{0.22} & \meanStd{96.58}{0.11} \\
    \bottomrule
    \end{tabular}
    
    \label{tab:4PICP}
\end{table*}

%% file: tab-4MPIW.tex
\begin{table*}[ht!]
\caption[Model evaluation using MPIW at different confidence intervals.]{Models are evaluated using MPIW at confidence intervals of 50\%, 60\%, 70\%, 80\%, 90\% and 98\%. Results are averaged over 10 independent runs, with standard deviations shown in light gray. The best-performing results are highlighted in bold (EvDNN excluded).}
    \centering
    \begin{tabular}{lrrrrrr}
    \toprule
     % $[\text{EUR}]$    & \glsxtrshort{MPIW}$_{50\%}$ & \glsxtrshort{MPIW}$_{60\%}$ & \glsxtrshort{MPIW}$_{70\%}$ & \glsxtrshort{MPIW}$_{80\%}$ & \glsxtrshort{MPIW}$_{90\%}$ & \glsxtrshort{MPIW}$_{98\%}$ \\
     $[\text{EUR}]$    & MPIW$_{50\%}$ & MPIW$_{60\%}$ & MPIW$_{70\%}$ & MPIW$_{80\%}$ & MPIW$_{90\%}$ & MPIW$_{98\%}$ \\
    \midrule
    \modelName{Naive-HS$_{train}$} & \meanStd{57.14}{0.00} & \meanStd{71.29}{0.00} & \meanStd{87.80}{0.00} & \meanStd{108.56}{0.00} & \meanStd{139.34}{0.00} & \meanStd{197.07}{0.00} \\
    \modelName{Naive-HS$_{val}$}   & \meanStd{49.94}{0.00} & \meanStd{62.32}{0.00} & \meanStd{76.74}{0.00} & \meanStd{94.89}{0.00}  & \meanStd{121.79}{0.00} & \meanStd{172.25}{0.00} \\
    \modelName{DDNN}               & \meanStd{24.10}{2.34} & \meanStd{30.07}{2.92} & \meanStd{37.03}{3.60} & \meanStd{45.78}{4.45}  & \meanStd{58.76}{5.71}  & \meanStd{83.11}{8.08} \\
    \modelName{Ens5}               & \meanStd{26.20}{3.95} & \meanStd{32.31}{3.48} & \meanStd{39.86}{3.73} & \meanStd{52.84}{9.92}  & \meanStd{75.12}{23.83} & \meanStd{103.55}{24.82} \\
    \modelName{Ens10}              & \meanStd{24.88}{0.91} & \meanStd{31.14}{1.18} & \meanStd{38.54}{1.52} & \meanStd{48.07}{2.07}  & \meanStd{62.90}{3.30}  & \meanStd{94.10}{7.87} \\
    \modelName{MCD10}              & \meanStd{24.41}{1.02} & \meanStd{30.55}{1.28} & \meanStd{37.77}{1.58} & \meanStd{47.00}{1.98}  & \meanStd{61.01}{2.59}  & \meanStd{88.59}{3.83} \\
    \modelName{MCD30}              & \meanStd{24.52}{1.00} & \meanStd{30.69}{1.25} & \meanStd{37.96}{1.55} & \meanStd{47.25}{1.92}  & \meanStd{61.39}{2.49}  & \meanStd{89.42}{3.62} \\
    \modelName{EvDNN}              & \meanStd{2.26}{0.30}  & \meanStd{2.87}{0.38}  & \meanStd{3.62}{0.49}  & \meanStd{4.64}{0.65}  & \meanStd{6.41}{0.94}   & \meanStd{11.08}{1.90} \\
    \modelName{LEAR-GARCH} & \meanStd{23.42}{0.00} & \meanStd{29.22}{0.00} & \meanStd{35.98}{0.00} & \meanStd{44.49}{0.00} & \meanStd{57.10}{0.00} & \meanStd{\textbf{80.76}}{0.00} \\
    \modelName{LEAR-QRA} & \meanStd{\textbf{17.88}}{0.00} & \meanStd{\textbf{23.15}}{0.00} & \meanStd{\textbf{29.63}}{0.00} & \meanStd{\textbf{39.18}}{0.00} & \meanStd{54.13}{0.00} & \meanStd{98.68}{0.00} \\
    \modelName{DDNN-CP} & \meanStd{24.64}{1.19} & \meanStd{31.27}{1.39} & \meanStd{39.51}{1.70} & \meanStd{50.41}{1.95} & \meanStd{66.66}{2.41} & \meanStd{100.51}{4.34} \\
    \modelName{LEAR-CP} & \meanStd{20.46}{0.00} & \meanStd{25.61}{0.00} & \meanStd{31.74}{0.00} & \meanStd{39.72}{0.00} & \meanStd{\textbf{52.84}}{0.00} & \meanStd{83.60}{0.00} \\
    \modelName{Ens10-CP} & \meanStd{23.08}{0.32} & \meanStd{29.53}{0.40} & \meanStd{37.60}{0.56} & \meanStd{48.40}{0.74} & \meanStd{64.03}{1.01} & \meanStd{96.02}{1.48} \\
    \modelName{MCD30-CP} & \meanStd{23.05}{0.71} & \meanStd{29.27}{0.77} & \meanStd{37.06}{0.87} & \meanStd{47.21}{0.93} & \meanStd{62.91}{0.83} & \meanStd{95.61}{1.16} \\
    \modelName{EvDNN-CP} & \meanStd{22.22}{1.36} & \meanStd{28.38}{1.59} & \meanStd{35.89}{1.96} & \meanStd{45.98}{2.35} & \meanStd{61.61}{2.83} & \meanStd{93.93}{2.79} \\
    \bottomrule
    \end{tabular}
    
    \label{tab:4MPIW}
\end{table*}

%% file: tab-hp_opt.tex
\begin{table*}[ht!]
\tiny
\caption{Hyperparameters and corresponding search ranges considered during tuning for each model type.}
    \centering
    \begin{tabular}{lrrrr}
    \toprule
                                            & LEAR(-GARCH/QRA)      & DDNN                  &MCD(10/30)             &EvDNN                  \\
    \midrule
        Learning rate                       & -                     & $[10^{-5},10^{-1}]$   & $[10^{-5},10^{-1}]$   &$[10^{-5},10^{-1}]$    \\
        $L_2$ regularization parameter      & -                     & $[10^{-5},10^{-1}]$   & $[10^{-5},10^{-1}]$   & $[10^{-5},10^{-1}]$   \\
        Dropout rate                        & -                     & -                     & $[0.01,0.9]$          & -                     \\
        Evidence regularization parameter   & -                     & -                     & -                     & $[10^{-4},1]$         \\
        LASSO penalty parameter      & $[10^{-5},10^{-1}]$   & -   & -  & -   \\
    \bottomrule
    \end{tabular}
\label{tab:hp_opt}
\end{table*}

%% file: tab-TProfit.tex
\begin{table*}[ht!]
\tiny
\caption[Total profit of models across different confidence intervals.]{Total profit of models is shown for confidence intervals of 50\%, 60\%, 70\%, 80\%, 90\% and 98\% using the quantile-based strategy. For comparison, the optimal bids strategy is also included. Results are averaged over 10 independent runs, with standard deviations shown in light gray. The best-performing results are highlighted in bold.}
\label{tab:4TProfit}
    \centering
    \begin{tabular}{lrrrrrr}
    \toprule
    Total profit     & PI$_{50\%}$ & PI$_{60\%}$ & PI$_{70\%}$ & PI$_{80\%}$ & PI$_{90\%}$ & PI$_{98\%}$ \\
    $[\text{EUR}]$ &&&&&& \\
    \midrule
    Optimal bids &\multicolumn{6}{c}{\meanStdh{32543}{0}}\\
    \midrule
    % \modelName{Naive-HS$_{train}$} & \meanStd{19252}{0}   & \meanStd{17341}{0}   & \meanStd{15875}{0}   & \meanStd{13348}{0}   & \meanStd{7037}{0}   & \meanStd{1462}{0}   \\
    \modelName{Naive-HS$_{val}$}   & \meanStd{19280}{0}   & \meanStd{17876}{0}   & \meanStd{15866}{0}   & \meanStd{12550}{0}   & \meanStd{6669}{0}   & \meanStd{1389}{0}   \\
    \modelName{DDNN}               & \meanStd{25941}{583} & \meanStd{25397}{900} & \meanStd{24120}{1458}& \meanStd{22067}{1757}& \meanStd{17891}{2592}& \meanStd{8491}{3551} \\
    % \modelName{Ens5}               & \meanStd{26414}{964} & \meanStd{25623}{1000}& \meanStd{24559}{1031}& \meanStd{22254}{1202}& \meanStd{17640}{1457}& \meanStd{6134}{1290} \\
    \modelName{Ens10}              & \meanStd{26695}{410} & \meanStd{25907}{496} & \meanStd{25014}{587} & \meanStd{22786}{951} & \meanStd{18175}{1373}& \meanStd{5586}{1864} \\
    % \modelName{MCD10}              & \meanStd{26940}{527} & \meanStd{26798}{485} & \meanStd{26125}{758} & \meanStd{23819}{887} & \meanStd{18188}{1624}& \meanStd{4666}{1614} \\
    \modelName{MCD30}              & \meanStd{26959}{645} & \meanStd{26754}{554} & \meanStd{26135}{848} & \meanStd{23735}{1035}& \meanStd{18346}{1644}& \meanStd{4217}{1441} \\
    % \modelName{EvDNN}              & \meanStd{25290}{1405}& \meanStd{25446}{1320}& \meanStd{25524}{1332}& \meanStd{25692}{1272}& \meanStd{25964}{1211}& \meanStd{\textbf{26539}}{1171} \\
    \modelName{LEAR-GARCH}         & \meanStd{27368}{0}   & \meanStd{27652}{0}   & \meanStd{27749}{0}   & \meanStd{26004}{0}   & \meanStd{23296}{0}   & \meanStd{16485}{0}   \\
    \modelName{LEAR-QRA}           & \meanStd{\textbf{28427}}{0} & \meanStd{\textbf{28122}}{0} & \meanStd{\textbf{28476}}{0} & \meanStd{\textbf{27766}}{0} & \meanStd{\textbf{25704}}{0} & \meanStd{10298}{0} \\
    \modelName{DDNN-CP}            & \meanStd{24887}{714} & \meanStd{23976}{873} & \meanStd{22281}{1093}& \meanStd{19585}{1121}& \meanStd{14386}{1304}& \meanStd{4699}{731} \\
    \modelName{LEAR-CP}            & \meanStd{26757}{0}   & \meanStd{26653}{0}   & \meanStd{26388}{0}   & \meanStd{26733}{0}   & \meanStd{26224}{0}   & \meanStd{\textbf{22108}}{0} \\
    \modelName{Ens10-CP}           & \meanStd{25679}{379} & \meanStd{24964}{380} & \meanStd{23279}{546} & \meanStd{21065}{413} & \meanStd{15846}{500} & \meanStd{5349}{322} \\
    \modelName{MCD30-CP}           & \meanStd{26111}{534} & \meanStd{25462}{556} & \meanStd{24299}{653} & \meanStd{21819}{957} & \meanStd{17212}{1315}& \meanStd{5601}{755} \\
    % \modelName{EvDNN-CP}           & \meanStd{26173}{1127}& \meanStd{25695}{1109}& \meanStd{23912}{1036}& \meanStd{20841}{1027}& \meanStd{15133}{798} & \meanStd{4898}{916} \\
    \bottomrule
  \end{tabular}
  
\end{table*}

%% file: tab-4exdp.tex
\newcommand{\modelNametwo}[1]{%
  \shortstack[r]{
  #1 \\ 
  \color{white}{0} \\ 
  \color{white}{0}
  }%
}

\begin{table*}[ht!]
\tiny
\caption[Expected number of days with profitable limit trades across different confidence intervals.]{The expected number of days with profitable limit trades (satisfying Equation \ref{eq:limit_profitable}, \ref{eq:limit_profitable_B2}, or \ref{eq:limit_profitable_B0}) is shown for confidence intervals of 50\%, 60\%, 70\%, 80\%, 90\% and 98\% using the quantile-based strategy. Results are averaged over 10 independent runs, with standard deviations shown in light gray. The best-performing results are highlighted in bold.}
\label{tab:4exdp}  
    \centering
    \begin{tabular}{lrrrrrr}
    \toprule
    \shortstack[l]{Expected number of \\days with profitable\\ limit trades}  & \modelNametwo{PI$_{50\%}$\\} & \modelNametwo{PI$_{60\%}$} & \modelNametwo{PI$_{70\%}$} & \modelNametwo{PI$_{80\%}$} & \modelNametwo{PI$_{90\%}$} & \modelNametwo{PI$_{98\%}$} \\
    $[\%]$ &&&&&& \\
    \midrule
    % \modelName{Naive-HS\(_{train}\)} & \meanStd{63.66}{0.00} & \meanStd{51.37}{0.00} & \meanStd{42.35}{0.00} & \meanStd{33.88}{0.00} & \meanStd{17.53}{0.00} & \meanStd{3.84}{0.00} \\
    \modelName{Naive-HS\(_{val}\)}   & \meanStd{65.30}{0.00} & \meanStd{56.28}{0.00} & \meanStd{43.99}{0.00} & \meanStd{31.97}{0.00} & \meanStd{16.44}{0.00} & \meanStd{4.11}{0.00} \\
    \modelName{DDNN}                 & \meanStd{91.39}{3.09} & \meanStd{84.87}{4.99} & \meanStd{75.35}{6.51} & \meanStd{64.41}{7.22} & \meanStd{47.40}{7.77} & \meanStd{21.01}{8.08} \\
    % \modelName{Ens5}                 & \meanStd{92.55}{2.39} & \meanStd{84.36}{3.16} & \meanStd{74.77}{3.14} & \meanStd{61.53}{3.97} & \meanStd{44.11}{4.09} & \meanStd{14.11}{3.28} \\
    \modelName{Ens10}                & \meanStd{92.50}{1.13} & \meanStd{84.60}{1.84} & \meanStd{75.64}{2.45} & \meanStd{63.01}{3.18} & \meanStd{45.37}{3.76} & \meanStd{12.96}{4.08} \\
    % \modelName{MCD10}                & \meanStd{94.01}{1.39} & \meanStd{89.15}{1.85} & \meanStd{82.00}{2.29} & \meanStd{69.40}{3.53} & \meanStd{47.37}{4.56} & \meanStd{11.34}{3.54} \\
    \modelName{MCD30}                & \meanStd{94.11}{1.11} & \meanStd{89.24}{1.35} & \meanStd{81.67}{2.36} & \meanStd{69.07}{4.23} & \meanStd{47.21}{4.79} & \meanStd{10.11}{3.06} \\
    % \modelName{EvDNN}                & \meanStd{\textbf{99.48}}{0.19} & \meanStd{\textbf{99.48}}{0.19} & \meanStd{\textbf{99.45}}{0.24} & \meanStd{\textbf{99.40}}{0.20} & \meanStd{\textbf{99.13}}{0.29} & \meanStd{\textbf{97.78}}{0.71} \\
    \modelName{LEAR-GARCH}           & \meanStd{92.90}{0.00} & \meanStd{91.26}{0.00} & \meanStd{84.70}{0.00} & \meanStd{76.23}{0.00} & \meanStd{61.48}{0.00} & \meanStd{37.43}{0.00} \\
    \modelName{LEAR-QRA}             & \meanStd{\textbf{95.36}}{0.00} & \meanStd{\textbf{91.80}}{0.00} & \meanStd{88.52}{0.00} & \meanStd{81.97}{0.00} & \meanStd{69.67}{0.00} & \meanStd{24.86}{0.00} \\
    \modelName{DDNN-CP}              & \meanStd{87.02}{1.80} & \meanStd{78.83}{2.68} & \meanStd{68.55}{3.47} & \meanStd{55.44}{3.57} & \meanStd{37.79}{3.02} & \meanStd{10.79}{1.72} \\
    \modelName{LEAR-CP}              & \meanStd{92.90}{0.00} & \meanStd{91.53}{0.00} & \meanStd{\textbf{89.62}}{0.00} & \meanStd{\textbf{86.89}}{0.00} & \meanStd{\textbf{82.51}}{0.00} & \meanStd{\textbf{60.66}}{0.00} \\
    \modelName{Ens10-CP}             & \meanStd{89.34}{0.59} & \meanStd{83.11}{0.78} & \meanStd{71.89}{1.37} & \meanStd{59.78}{1.11} & \meanStd{41.12}{1.11} & \meanStd{12.62}{0.70} \\
    \modelName{MCD30-CP}             & \meanStd{91.01}{0.78} & \meanStd{85.05}{1.03} & \meanStd{76.28}{1.38} & \meanStd{63.33}{2.84} & \meanStd{44.86}{2.87} & \meanStd{12.76}{1.67} \\
    % \modelName{EvDNN-CP}             & \meanStd{90.19}{0.93} & \meanStd{83.61}{1.66} & \meanStd{72.62}{1.68} & \meanStd{58.83}{2.55} & \meanStd{40.49}{2.56} & \meanStd{11.53}{2.37} \\
    \bottomrule
  \end{tabular}
  
\end{table*}

%% file: tab-4ptp.tex
\begin{table*}[ht!]
\tiny
\caption[Per-transaction profit of models across different confidence intervals.]{Per-transaction profit of the models is shown for confidence intervals of 50\%, 60\%, 70\%, 80\%, 90\% and 98\% using the quantile-based strategy. For comparison, the optimal bids strategy is also included. Results are averaged over 10 independent runs, with standard deviations shown in light gray. The best-performing results are highlighted in bold.}
\label{tab:4ptp}  
    \centering
    \begin{tabular}{lrrrrrr}
    \toprule
    Per-transaction profit      & PI$_{50\%}$ & PI$_{60\%}$ & PI$_{70\%}$ & PI$_{80\%}$ & PI$_{90\%}$ & PI$_{98\%}$ \\
    $[\text{EUR}]$ &&&&&& \\
    \midrule
    Optimal bids &\multicolumn{6}{c}{\meanStdh{44.46}{0.00}}\\
    \midrule
    % \modelName{Naive-HS$_{train}$} & \meanStd{\textbf{45.19}}{0.00} & \meanStd{\textbf{49.26}}{0.00} & \meanStd{\textbf{52.57}}{0.00} & \meanStd{53.82}{0.00} & \meanStd{55.85}{0.00} & \meanStd{52.23}{0.00} \\
    \modelName{Naive-HS$_{val}$}   & \meanStd{\textbf{44.43}}{0.00} & \meanStd{\textbf{46.55}}{0.00} & \meanStd{\textbf{50.21}}{0.00} & \meanStd{\textbf{54.56}}{0.00} & \meanStd{\textbf{59.54}}{0.00} & \meanStd{46.29}{0.00} \\
    \modelName{DDNN}               & \meanStd{40.80}{1.12} & \meanStd{42.70}{1.48} & \meanStd{45.50}{1.77} & \meanStd{48.45}{1.80} & \meanStd{52.92}{1.91} & \meanStd{55.04}{3.58} \\
    % \modelName{Ens5}               & \meanStd{40.68}{0.81} & \meanStd{42.81}{0.83} & \meanStd{45.84}{0.85} & \meanStd{50.19}{1.36} & \meanStd{55.14}{1.25} & \meanStd{\textbf{59.77}}{1.93} \\
    \modelName{Ens10}              & \meanStd{41.21}{0.45} & \meanStd{43.42}{0.58} & \meanStd{46.32}{0.70} & \meanStd{50.18}{1.17} & \meanStd{55.18}{0.68} & \meanStd{58.71}{2.31} \\
    % \modelName{MCD10}              & \meanStd{41.00}{1.01} & \meanStd{42.41}{0.95} & \meanStd{44.49}{1.28} & \meanStd{47.57}{1.23} & \meanStd{52.89}{1.43} & \meanStd{55.73}{3.93} \\
    \modelName{MCD30}              & \meanStd{40.81}{1.04} & \meanStd{42.14}{1.10} & \meanStd{44.58}{1.29} & \meanStd{47.43}{1.30} & \meanStd{53.62}{1.14} & \meanStd{56.35}{4.04} \\
    % \modelName{EvDNN}              & \meanStd{41.53}{1.02} & \meanStd{41.62}{1.11} & \meanStd{41.49}{1.19} & \meanStd{41.33}{1.12} & \meanStd{41.28}{1.15} & \meanStd{41.34}{0.93} \\
    \modelName{LEAR-GARCH}         & \meanStd{43.44}{0.00} & \meanStd{43.61}{0.00} & \meanStd{45.79}{0.00} & \meanStd{46.94}{0.00} & \meanStd{51.77}{0.00} & \meanStd{60.16}{0.00} \\
    \modelName{LEAR-QRA}           & \meanStd{42.94}{0.00} & \meanStd{43.94}{0.00} & \meanStd{45.20}{0.00} & \meanStd{47.54}{0.00} & \meanStd{50.80}{0.00} & \meanStd{56.58}{0.00} \\
    \modelName{DDNN-CP}            & \meanStd{41.88}{0.99} & \meanStd{43.83}{1.02} & \meanStd{46.22}{0.99} & \meanStd{49.78}{0.97} & \meanStd{52.82}{1.75} & \meanStd{60.29}{3.70} \\
    \modelName{LEAR-CP}            & \meanStd{42.07}{0.00} & \meanStd{41.78}{0.00} & \meanStd{42.02}{0.00} & \meanStd{42.98}{0.00} & \meanStd{44.00}{0.00} & \meanStd{50.02}{0.00} \\
    \modelName{Ens10-CP}           & \meanStd{41.65}{0.43} & \meanStd{42.73}{0.57} & \meanStd{45.79}{0.65} & \meanStd{49.17}{0.68} & \meanStd{53.25}{0.98} & \meanStd{58.02}{1.00} \\
    \modelName{MCD30-CP}           & \meanStd{41.31}{0.98} & \meanStd{42.48}{1.21} & \meanStd{44.73}{0.86} & \meanStd{47.88}{0.75} & \meanStd{52.90}{1.10} & \meanStd{\textbf{60.45}}{2.63} \\
    % \modelName{EvDNN-CP}           & \meanStd{42.77}{1.52} & \meanStd{44.55}{1.33} & \meanStd{47.23}{1.67} & \meanStd{49.81}{1.49} & \meanStd{51.86}{1.16} & \meanStd{58.28}{1.74} \\
    \bottomrule
  \end{tabular}
  
\end{table*}

%% file: tab-dbvalues.tex
\newcolumntype{Y}{>{\raggedleft\arraybackslash}X}

\begin{table*}[ht!]
\tiny
\caption{Pairwise DM statistics $\widehat{DM}_{i,j}$ (row model $i$ vs.\ column model $j$). Negative values favor the row model, positive values favor the column model. Results are averaged over 10 independent runs, yielding 100 comparisons per model pair across runs.}
\label{tab:db}
\centering
\label{tab:dm-matrix}
% \resizebox{\textwidth}{!}{%
% \begin{tabular}{lrrrrrrrrrr}
\begin{tabularx}{\textwidth}{l*{10}{Y}}
\toprule
 & Naive-HS$_{val}$ & DDNN & Ens10 & MCD30 & LEAR-GARCH & LEAR-QRA & DDNN-CP & LEAR-CP & Ens10-CP & MCD30-CP \\
\midrule
\modelName{Naive-HS$_{val}$} & \modelName{--} & \meanStd{6.273}{0.660} & \meanStd{6.984}{0.157} & \meanStd{7.387}{0.356} & \meanStd{10.619}{0.000} & \meanStd{10.514}{0.000} & \meanStd{6.433}{0.638} & \meanStd{10.235}{0.000} & \meanStd{6.898}{0.193} & \meanStd{7.232}{0.319} \\
\modelName{DDNN} & \meanStd{-6.273}{0.660} & \modelName{--} & \meanStd{2.432}{2.309} & \meanStd{3.087}{1.868} & \meanStd{5.763}{0.712} & \meanStd{5.781}{0.692} & \meanStd{0.495}{2.370} & \meanStd{5.400}{0.703} & \meanStd{2.037}{2.179} & \meanStd{2.650}{1.801} \\
\modelName{Ens10} & \meanStd{-6.984}{0.157} & \meanStd{-2.432}{2.309} & \modelName{--} & \meanStd{1.748}{1.396} & \meanStd{5.150}{0.231} & \meanStd{5.184}{0.240} & \meanStd{-1.778}{2.084} & \meanStd{4.784}{0.225} & \meanStd{-0.299}{1.419} & \meanStd{1.201}{1.244} \\
\modelName{MCD30} & \meanStd{-7.387}{0.356} & \meanStd{-3.087}{1.868} & \meanStd{-1.748}{1.396} & \modelName{--} & \meanStd{4.683}{0.361} & \meanStd{4.692}{0.388} & \meanStd{-2.700}{1.782} & \meanStd{4.313}{0.358} & \meanStd{-1.856}{1.312} & \meanStd{-0.438}{1.555} \\
\modelName{LEAR-GARCH} & \meanStd{-10.619}{0.000} & \meanStd{-5.763}{0.712} & \meanStd{-5.150}{0.231} & \meanStd{-4.683}{0.361} & \modelName{--} & \meanStd{-0.072}{0.000} & \meanStd{-5.731}{0.658} & \meanStd{-0.049}{0.000} & \meanStd{-5.264}{0.219} & \meanStd{-4.815}{0.340} \\
\modelName{LEAR-QRA} & \meanStd{-10.514}{0.000} & \meanStd{-5.781}{0.692} & \meanStd{-5.184}{0.240} & \meanStd{-4.692}{0.388} & \meanStd{0.072}{0.000} & \modelName{--} & \meanStd{-5.752}{0.636} & \meanStd{0.039}{0.000} & \meanStd{-5.305}{0.227} & \meanStd{-4.839}{0.377} \\
\modelName{DDNN-CP} & \meanStd{-6.433}{0.638} & \meanStd{-0.495}{2.370} & \meanStd{1.778}{2.084} & \meanStd{2.700}{1.782} & \meanStd{5.731}{0.658} & \meanStd{5.752}{0.636} & \modelName{--} & \meanStd{5.374}{0.658} & \meanStd{1.656}{2.251} & \meanStd{2.431}{1.816} \\
\modelName{LEAR-CP} & \meanStd{-10.235}{0.000} & \meanStd{-5.400}{0.703} & \meanStd{-4.784}{0.225} & \meanStd{-4.313}{0.358} & \meanStd{0.049}{0.000} & \meanStd{-0.039}{0.000} & \meanStd{-5.374}{0.658} & \modelName{--} & \meanStd{-4.921}{0.224} & \meanStd{-4.474}{0.340} \\
\modelName{Ens10-CP} & \meanStd{-6.898}{0.193} & \meanStd{-2.037}{2.179} & \meanStd{0.299}{1.419} & \meanStd{1.856}{1.312} & \meanStd{5.264}{0.219} & \meanStd{5.305}{0.227} & \meanStd{-1.656}{2.251} & \meanStd{4.921}{0.224} & \modelName{--} & \meanStd{1.528}{1.336} \\
\modelName{MCD30-CP} & \meanStd{-7.232}{0.319} & \meanStd{-2.650}{1.801} & \meanStd{-1.201}{1.244} & \meanStd{0.438}{1.555} & \meanStd{4.815}{0.340} & \meanStd{4.839}{0.377} & \meanStd{-2.431}{1.816} & \meanStd{4.474}{0.340} & \meanStd{-1.528}{1.336} & \modelName{--} \\
\bottomrule
% \end{tabular}
\end{tabularx}
% }
\end{table*}

%% file: tab-p-values.tex
\begin{table*}[t]
\tiny
\caption{Pairwise DM \emph{p}-values $p_{i,j}$ (row model $i$ vs.\ column model $j$). Results are averaged over 10 independent runs, yielding 100 comparisons per model pair across runs. For notational convenience, we write $p = 0.000$ whenever $p<0.001$.}
\label{tab:dm-pvals}
\centering
\begin{tabularx}{\textwidth}{l*{10}{Y}}
\toprule
 & Naive-HS$_{val}$ & DDNN & Ens10 & MCD30 & LEAR-GARCH & LEAR-QRA & DDNN-CP & LEAR-CP & Ens10-CP & MCD30-CP \\
\midrule
\modelName{Naive-HS$_{val}$} & \modelName{--} & \meanStd{1.000}{0.000} & \meanStd{1.000}{0.000} & \meanStd{1.000}{0.000} & \meanStd{1.000}{0.000} & \meanStd{1.000}{0.000} & \meanStd{1.000}{0.000} & \meanStd{1.000}{0.000} & \meanStd{1.000}{0.000} & \meanStd{1.000}{0.000} \\
\modelName{DDNN}            & \meanStd{0.000}{0.000} & \modelName{--} & \meanStd{0.808}{0.299} & \meanStd{0.930}{0.181} & \meanStd{1.000}{0.000} & \meanStd{1.000}{0.000} & \meanStd{0.578}{0.413} & \meanStd{1.000}{0.000} & \meanStd{0.776}{0.312} & \meanStd{0.903}{0.194} \\
\modelName{Ens10}           & \meanStd{0.000}{0.000} & \meanStd{0.192}{0.299} & \modelName{--} & \meanStd{0.870}{0.278} & \meanStd{1.000}{0.000} & \meanStd{1.000}{0.000} & \meanStd{0.243}{0.324} & \meanStd{1.000}{0.000} & \meanStd{0.422}{0.349} & \meanStd{0.809}{0.277} \\
\modelName{MCD30}           & \meanStd{0.000}{0.000} & \meanStd{0.070}{0.181} & \meanStd{0.130}{0.278} & \modelName{--} & \meanStd{1.000}{0.000} & \meanStd{1.000}{0.000} & \meanStd{0.090}{0.203} & \meanStd{1.000}{0.000} & \meanStd{0.118}{0.268} & \meanStd{0.376}{0.289} \\
\modelName{LEAR-GARCH}      & \meanStd{0.000}{0.000} & \meanStd{0.000}{0.000} & \meanStd{0.000}{0.000} & \meanStd{0.000}{0.000} & \modelName{--} & \meanStd{0.471}{0.000} & \meanStd{0.000}{0.000} & \meanStd{0.480}{0.000} & \meanStd{0.000}{0.000} & \meanStd{0.000}{0.000} \\
\modelName{LEAR-QRA}        & \meanStd{0.000}{0.000} & \meanStd{0.000}{0.000} & \meanStd{0.000}{0.000} & \meanStd{0.000}{0.000} & \meanStd{0.529}{0.000} & \modelName{--} & \meanStd{0.000}{0.000} & \meanStd{0.516}{0.000} & \meanStd{0.000}{0.000} & \meanStd{0.000}{0.000} \\
\modelName{DDNN-CP}         & \meanStd{0.000}{0.000} & \meanStd{0.422}{0.413} & \meanStd{0.757}{0.324} & \meanStd{0.910}{0.203} & \meanStd{1.000}{0.000} & \meanStd{1.000}{0.000} & \modelName{--} & \meanStd{1.000}{0.000} & \meanStd{0.722}{0.348} & \meanStd{0.882}{0.220} \\
\modelName{LEAR-CP}         & \meanStd{0.000}{0.000} & \meanStd{0.000}{0.000} & \meanStd{0.000}{0.000} & \meanStd{0.000}{0.000} & \meanStd{0.520}{0.000} & \meanStd{0.484}{0.000} & \meanStd{0.000}{0.000} & \modelName{--} & \meanStd{0.000}{0.000} & \meanStd{0.000}{0.000} \\
\modelName{Ens10-CP}        & \meanStd{0.000}{0.000} & \meanStd{0.224}{0.312} & \meanStd{0.578}{0.349} & \meanStd{0.882}{0.268} & \meanStd{1.000}{0.000} & \meanStd{1.000}{0.000} & \meanStd{0.278}{0.348} & \meanStd{1.000}{0.000} & \modelName{--} & \meanStd{0.846}{0.274} \\
\modelName{MCD30-CP}        & \meanStd{0.000}{0.000} & \meanStd{0.097}{0.194} & \meanStd{0.191}{0.277} & \meanStd{0.624}{0.289} & \meanStd{1.000}{0.000} & \meanStd{1.000}{0.000} & \meanStd{0.118}{0.220} & \meanStd{1.000}{0.000} & \meanStd{0.154}{0.274} & \modelName{--} \\
\bottomrule
\end{tabularx}
\end{table*}